\documentclass[12pt]{article}
\usepackage{fullpage,graphicx,psfrag,amsmath,amsfonts,verbatim}
\usepackage{mystyle}
\usepackage[small,bf]{caption}
\usepackage{xcolor}
\usepackage{amsthm}
\usepackage{hyperref}
\usepackage{upgreek}
\hypersetup{
    colorlinks=true,
    linkcolor=blue,
    filecolor=magenta,
    urlcolor=cyan,
    pdftitle={Learning from Samples: Inverse Problems over Measures},
    pdfpagemode=UseOutlines,
    }

\usepackage{amssymb}
\usepackage[]{mdframed}
\newtheorem*{thm*}{Theorem}
\newtheorem*{rem*}{Remark}
\newtheorem*{prop*}{Proposition}
\newtheorem*{lemma*}{Lemma}
\usepackage[left=2.4cm, right=2.4cm]{geometry}

\usepackage{mathtools}

\renewcommand{\argmin}{\mathop{\mathrm{argmin}}}

\renewcommand{\argmax}{\mathop{\mathrm{argmax}}}

\newcommand{\KL}{\mathrm{KL}}

\newtheorem{innerassumption}{Assumption}
\newenvironment{assumption}
  {\innerassumption}
  {\endinnerassumption}

\author{
Francisco Andrade
\thanks{INRIA (PreMeDICaL \& HeKA), 
\texttt{francisco.de-lima-andrade@inria.fr}}
\and
Gabriel Peyr\'e
\thanks{CNRS and ENS, PSL Universit\'e,
\texttt{gabriel.peyre@ens.fr}}
\and
Clarice Poon \thanks{
Mathematics Institute, University of Warwick,
\texttt{clarice.poon@warwick.ac.uk}}
}

\title{Learning from Samples: Inverse Problems over Measures}

\date{}
\begin{document}
\maketitle

\begin{abstract}
We study inverse problems where an unknown potential is observed only through
samples from the measure it induces by a convex variational principle. Such
problems arise in learning costs, energies, and dynamics from distributional
data, but the associated forward solution map is typically nonlinear and
implicit. We show that its optimality gap nevertheless yields convex empirical
objectives for finite-dimensional potential classes, and we introduce sharpened
Fenchel--Young losses that add a data-dependent discrepancy inside the forward
problem. This keeps the estimator calibrated while improving the local geometry
of the loss. Our main stability theorem separates the inverse error analysis
into measurement error, forward perturbation, and empirical curvature. We
instantiate this principle for inverse entropic unbalanced optimal transport and
for inverse Jordan--Kinderlehrer--Otto (JKO) learning from independent snapshot
samples, obtaining high-probability parameter recovery bounds. JKO schemes
discretize Wasserstein gradient flows through a sequence of variational
problems over measures, making them a natural language for population dynamics
observed through snapshots. In this JKO case, the sharpened objective reduces
to an unbalanced transport problem, which also clarifies the connection between
variational gap losses and quadratic iJKO\(^\star\) surrogates.
Numerical experiments illustrate the conditioning effect of sharpening and its
benefits for sparse inverse-gradient-flow recovery.
\end{abstract}

\tableofcontents

\section{Introduction}\label{sec:introduction}

Modern machine learning increasingly works with distributions as data objects:
matching markets and recommendation systems produce observed couplings
\cite{galichon2022cupid,dupuy2019estimating,li2019learning}, generative and
energy-based models define samples through latent energies
\cite{blondel2022learning,genevay2018learning,de2021diffusion}, and biological
or physical systems are often observed only through population snapshots
\cite{schiebinger2019optimal,lavenant2021towards,bunne2022proximal}. In
these settings, one wants to learn the hidden cost, potential, or dynamical law
that generated the samples. This paper develops a convex
optimality-gap methodology for such inverse variational problems over measures,
introduces a sharpened loss construction to improve statistical stability, and
proves finite-sample recovery guarantees for inverse unbalanced transport and
inverse JKO learning.

The main difficulty is that the parameter acts indirectly: the forward problem
is convex in the measure, but the parameter-to-measure map is nonlinear and
implicit. Instead of comparing empirical and predicted measures through a
possibly nonconvex discrepancy, we use the optimality gap of the forward problem
as the inverse loss. For affine potential classes this gives a convex
Fenchel--Young objective. We then sharpen the gap by adding a data-dependent
discrepancy inside the forward problem, which preserves calibration while
improving local curvature. The theory reduces recovery to measurement
concentration, forward stability, and empirical curvature, and we verify these
ingredients for inverse unbalanced transport and inverse JKO from independent
snapshots.

\subsection{Previous works}\label{sec:previous-works}

\paragraph{Inverse optimization.}
Inverse optimization asks for parameters of an optimization problem from
observed optimal or nearly optimal decisions. Early work of
Zhang and Liu~\cite{zhang1996calculating} and Ahuja and
Orlin~\cite{ahuja2001inverse} showed that several inverse linear and network
optimization problems admit tractable reformulations; see also the survey of
Heuberger~\cite{heuberger2004inverse} for the combinatorial inverse-optimization
literature. More recent data-driven formulations replace exact optimality by
losses based on KKT residuals, duality gaps, or distances to optimality sets.
Keshavarz, Wang, and Boyd~\cite{keshavarz2011imputing} estimate convex
objectives from nearly optimal decisions, Bertsimas, Gupta, and
Paschalidis~\cite{bertsimas2015data} use inverse optimization for estimation in
equilibrium models, Aswani, Shen, and Siddiq~\cite{aswani2018inverse} study
statistical consistency under noisy observations, and Chan, Lee, and
Terekhov~\cite{chan2018inverse} develop goodness-of-fit criteria for inverse
linear optimization. The recent survey of Chan, Mahmood, and
Zhu~\cite{chan2023inverse} gives a broader account of classical and
data-driven inverse optimization. This line of work is closest in spirit to our
formulation: observations are not generated by an explicit regression map, but
by optimality conditions. The measure-valued setting considered here adds two
difficulties absent from most classical inverse-optimization models: the
decision variable is a distribution, and the observations are empirical samples
rather than exact optimizers.

\paragraph{Fenchel--Young and Fitzpatrick losses.}
Fenchel--Young losses were introduced in machine learning as convex losses
generated by regularized prediction maps. Blondel et
al.~\cite{blondel2020learning} developed their differentiability and convexity
properties for structured prediction, where the forward map is finite
dimensional and usually evaluated exactly. Blondel et
al.~\cite{blondel2022learning} later generalized this viewpoint to energy
networks, emphasizing the role of convex duality in learning implicit
prediction rules. Carlier~\cite{carlier2023fenchel} studied quantitative
Fenchel--Young inequalities with remainders, which is closely related to asking
when a gap controls distance to optimality. Bauschke, Singh, and
Wang~\cite{bauschke2022carlier} clarified this inequality from the perspective
of monotone operator theory. Fitzpatrick functions provide another convex
representation of monotone inclusions, beginning with
Fitzpatrick~\cite{fitzpatrick1988representing}; recent learning formulations
based on Fitzpatrick losses, such as Rakotomandimby et
al.~\cite{rakotomandimby2024learning}, are therefore conceptually close to
gap-based inverse losses. Our contribution is to adapt this convex-duality
viewpoint to variational problems over measures and to introduce sharpening as
a data-dependent way to improve curvature.

\paragraph{Measure matching and kernel losses.}
A standard alternative to gap minimization is to compare the observed empirical
measure with the measure predicted by the forward model. Maximum mean
discrepancy losses are a prominent example: the kernel embedding framework of
Smola et al.~\cite{smola2007hilbert} and the two-sample testing theory of
Gretton et al.~\cite{gretton2012kernel} provide Hilbertian distances between
probability measures that are easy to estimate from samples; see
Muandet et al.~\cite{muandet2017kernel} for a review of kernel mean embeddings.
MMD criteria are used not only for testing but also as estimation objectives.
For instance, generative moment matching networks~\cite{li2015generative},
MMD nets~\cite{dziugaite2015training}, and MMD-GAN variants
\cite{li2017mmd,binkowski2018demystifying} fit neural generators by matching
generated and observed samples. In statistics, Briol et
al.~\cite{briol2019statistical} study MMD minimum-distance estimators for
simulator-based generative models. Such losses are widely useful, but in the
present inverse problems they are applied after the nonlinear forward solution
map. This typically produces a nonlinear objective in the unknown potential.
Section~\ref{sec:losses} makes this comparison explicit by showing that
MMD-type objectives can be interpreted as squared, preconditioned gradient norms
of the Fenchel--Young objective.

\paragraph{Entropic and unbalanced optimal transport.}
Optimal transport gives a variational model for couplings between probability
measures; Santambrogio~\cite{santambrogio2015optimal} provides a standard
reference for its mathematical foundations. Entropic regularization connects
transport to Schr\"odinger bridge problems, as surveyed by
L\'eonard~\cite{leonard2013survey}, and makes transport computationally
tractable through Sinkhorn scaling. The classical matrix scaling algorithm goes
back to Sinkhorn~\cite{Sinkhorn64}, while Cuturi~\cite{CuturiSinkhorn} made
entropic OT a practical tool in machine learning. Statistical properties of
entropic OT have been studied from several angles: Genevay et
al.~\cite{genevay2019sample} analyze Sinkhorn divergences, Mena and
Niles-Weed~\cite{mena2019statistical} study statistical bounds and limit
theorems, and Rigollet and Stromme~\cite{rigollet2022sample} give sample
complexity results for entropic transport. Unbalanced optimal transport
relaxes the hard marginal constraints of classical OT and is therefore better
suited to matching measures with unequal masses or partial observations.
Liero, Mielke, and Savar\'e~\cite{liero2018optimal} introduced the
entropy-transport formulation and the Hellinger--Kantorovich geometry. Chizat
et al.~\cite{chizat2018unbalanced} developed dynamic and Kantorovich
formulations that clarified the geometry of mass variation. S\'ejourn\'e,
Peyr\'e, and Vialard~\cite{sejourne2023unbalanced} provide a modern account of
theory and numerical methods. We rely on this forward statistical and
geometric literature as ingredients in an inverse stability argument. In
particular, unbalanced marginal penalties can remove some of the additive
ambiguities that appear in balanced inverse OT, but they also require
controlling the perturbation of empirical reference measures.

\paragraph{Inverse optimal transport and learned costs.}
Inverse optimal transport seeks to recover a ground cost from observed optimal
couplings. In economics and matching, Galichon and
Salani\'e~\cite{galichon2022cupid} used entropic matching models to connect
observed pairings with latent affinities, and Dupuy, Galichon, and
Sun~\cite{dupuy2019estimating} developed estimation of affinity matrices under
low-rank constraints. H\"utter, Mao, Rigollet, and
Robeva~\cite{hutter2020monge} studied estimation under the Monge matrix
constraint, showing how discrete transport structure can be used as a
statistical prior for recovering ordered affinity or cost arrays. Carlier,
Dupuy, Galichon, and
Sun~\cite{carlier2023sista} proposed sparse inverse transport formulations,
showing how convex regularization can select interpretable costs. Learning
costs or affinities is also motivated by applications where observed flows are
modeled as optimal or entropy-regularized allocations. Wilson's statistical
theory of spatial distributions~\cite{wilson1967statistical} already connected
entropy and transportation flows, while the gravity-model literature summarized
by Erlander and Stewart~\cite{erlander1990gravity} provides a classical
baseline for mobility and trade. Modern mobility modeling, surveyed by
Barbosa et al.~\cite{barbosa2018human}, often requires learning interaction
costs from aggregate movement data. Recent data-driven gravity models, such as
Simini et al.~\cite{simini2021deep} and Cabanas-Tirapu et
al.~\cite{cabanas2025human}, illustrate the continuing importance of
interpretable flow models.

In machine learning, Li et al.~\cite{li2019learning} used inverse OT for
learning to match, while Liu et al.~\cite{liu2019learning} considered learning
costs from subset correspondences. Ma et al.~\cite{ma2020learning} studied
parametric cost learning for OT, and Chiu, Wang, and
Shafto~\cite{chiu2022discrete} developed a discrete probabilistic inverse OT
model. More recent work uses inverse entropic OT for semi-supervised learning
and likelihood maximization, as in Persiianov et
al.~\cite{persiianov2024inverse}. In single-cell genomics, Samaran, Peyr\'e,
and Cantini~\cite{samaran2026champollion} introduce CHAMPOLLION, which uses
inverse optimal transport to learn an interpretable cross-modal metric from
paired cells for robust multi-omics integration. These applications motivate
inverse unbalanced formulations, because observed flows need not conserve the
prescribed source and target masses exactly. Our earlier
work~\cite{andradesparsistency} established sparsistency for inverse optimal
transport; the present paper moves from balanced inverse OT to unbalanced
transport and from structural recovery toward sample-level stability and
curvature of sharpened gap losses.

\paragraph{Distributional dynamics, JKO schemes, and generative bridges.}
Many inverse problems for dynamical systems assume trajectory observations.
Lu et al.~\cite{lu2019nonparametric} give a nonparametric theory for learning
interaction laws in agent systems from trajectory data. In contrast, many
biological and social datasets provide only samples from marginal
distributions at a few times. In single-cell genomics, Schiebinger et
al.~\cite{schiebinger2019optimal} used optimal transport to connect
time-stamped gene-expression snapshots and identify developmental trajectories
in reprogramming. Sha, Qiu, Zhou, and Nie~\cite{sha2023reconstructing}
reconstruct growth and dynamic trajectories from single-cell transcriptomics,
highlighting that snapshot dynamics often combine transport with changing
population mass. Lavenant, Zhang, Kim, and
Schiebinger~\cite{lavenant2021towards} study mathematical foundations for
trajectory inference from such distributional observations. More recently,
inverse-transport ideas have also appeared in multi-omics integration:
CHAMPOLLION~\cite{samaran2026champollion} learns a cross-modal cost from paired
cells and then uses it to align unpaired molecular modalities. This distinction
is central for inverse JKO:\@ we do not observe paired paths, but independent
samples from successive snapshots. JKO-type variational schemes model a time
step of a Wasserstein gradient flow as an optimization problem over measures.
Bunne et al.~\cite{bunne2022proximal} introduced JKO-Net, a bilevel approach
for learning population dynamics through proximal optimal transport. Terpin et
al.~\cite{terpin2024learning} proposed a quadratic inverse-JKO loss derived
from first-order optimality conditions, giving a computationally attractive
surrogate when the transport map or potential is estimated first. Zhang, Li,
and Zhou~\cite{zhang2024learning} study stochastic dynamics from snapshots
through regularized unbalanced OT, highlighting the relevance of mass variation
in dynamic settings. Action matching methods such as Neklyudov et
al.~\cite{neklyudov2023action} also learn dynamics from samples by matching
local transport or stochastic-action information.

Schr\"odinger bridge methods are closely related to entropic optimal transport
and have become important in generative modeling. De Bortoli et
al.~\cite{de2021diffusion} connected diffusion Schr\"odinger bridges with
score-based generative modeling. Somnath et al.~\cite{somnath2023aligned}
developed aligned diffusion Schr\"odinger bridges, while Shi et
al.~\cite{shi2024diffusion} proposed diffusion Schr\"odinger bridge matching.
Unbalanced variants, such as Pariset et al.~\cite{pariset2023unbalanced}, allow
for creation and destruction of mass. These works learn trajectories, bridges,
or stochastic interpolations between distributions; our focus is complementary,
namely recovering the potential or cost that appears in a variational step from
samples of its optimizer. Our approach keeps the JKO variational problem inside
a convex gap loss and shows that the quadratic iJKO\(^\star\) loss arises as a
high-sharpening local approximation.

\paragraph{Structured recovery and model identification.}
Several inverse problems seek not only parameter consistency but also recovery
of structure such as sparsity or low rank. The Lasso of
Tibshirani~\cite{tibshirani1996regression} and the sharp support-recovery
analysis of Wainwright~\cite{wainwright2009sharp} are canonical references for
sparse estimation. Nuclear-norm methods for low-rank recovery, such as Recht,
Fazel, and Parrilo~\cite{recht2010guaranteed} and Negahban and
Wainwright~\cite{negahban2011estimation}, play an analogous role for matrix
parameters. Vaiter et al.~\cite{vaiter2015model} and Fadili, Malick, and
Peyr\'e~\cite{fadili2018sensitivity} provide a variational framework for model
identification with partly smooth or mirror-stratifiable regularizers. This is
not the focus of the present paper: sparsity and support recovery for inverse
optimal transport are covered in our previous work on
sparsistency~\cite{andradesparsistency}. Here we only include sparsity-inducing
regularization in the numerical section, while the main theorems focus on
stability and finite-sample recovery for sharpened gap minimization.

\subsection{Contributions}

The results below are organized so that the modeling construction, the loss
design, the stability theory, and the numerical evidence can be read
independently while still feeding into one another. Each item points to the
section where the corresponding construction or theorem is developed.

\begin{itemize}
    \item \textbf{Inverse variational problems over measures.}
    Section~\ref{sec:inverse-problems-measures} formulates the statistical
    inverse problem for potentials observed through samples of an implicit
    variational optimizer. It also fixes the empirical-measure convention and
    presents the main examples used throughout the paper: likelihood-type
    density estimation, inverse optimal transport, inverse unbalanced optimal
    transport, and inverse JKO learning.

    \item \textbf{Gap losses and sharpening.}
    Section~\ref{sec:losses} introduces Fenchel--Young gap losses for inverse
    variational problems and the sharpened Fenchel--Young construction. The
    basic properties of the sharpened loss, including convexity,
    non-negativity, calibration, differentiability, and monotonicity in the
    sharpening term, are collected in Proposition~\ref{prop:key_properties}.
    The same section relates gap losses to MMD objectives and proves the
    high-sharpening connection with quadratic iJKO\(^\star\) losses in
    Proposition~\ref{prop:r_infty_ijko}.

    \item \textbf{Stability and finite-sample theory.}
    Section~\ref{sec:stability} proves the abstract stability theorem,
    Theorem~\ref{thm:stability}, which reduces inverse recovery to measurement
    concentration, forward stability, and local curvature. The inverse
    unbalanced OT application is developed in Section~\ref{sec:stability-iuot}:
    Theorem~\ref{thm:iuot-stability} gives the parameter recovery bound, using
    the forward sample-complexity estimate of
    Theorem~\ref{thm:sample-complexity-uot} and the local curvature result of
    Proposition~\ref{prop:strongConvexity}. The inverse JKO application is
    developed in Section~\ref{sec:stability-ijko}: the recovery guarantee is
    Theorem~\ref{thm:ijko-stability}, with the reduction to unbalanced
    transport quantified by Propositions~\ref{prop:ijko-forward-stability}
    and~\ref{prop:ijko-curvature}.

    \item \textbf{Numerical experiments.}
    Section~\ref{sec:numerics} illustrates the conditioning effect of
    sharpening in density estimation and the sparse-recovery behavior of the
    sharpened inverse-gradient-flow loss, including the improved optimization
    geometry in the precision-estimation example and support-recovery
    comparisons with the quadratic iJKO\(^\star\) baseline.
\end{itemize}

\section{Inverse problems over measures}\label{sec:inverse-problems-measures}

This section fixes the common inverse-problem template used throughout the
paper. The goal is to make precise what is observed, what is parametrized, and
where the implicit forward variational problem enters the statistical analysis.

Let \(\mathcal Z\) be a complete separable metric space, and let
\(\Mm_+(\Zz)\) denote the space of finite non-negative measures on \(\Zz\).
Let \(\Omega:\Mm_+(\Zz)\to \overline{\RR}\) be a proper functional, where
\(\overline{\RR}=\RR\cup\{+\infty\}\).

For a potential \(V\in\Cc(\Zz)\), we define the associated forward solution map by
\begin{equation}\label{eq:forward}
\mu_\Omega[V]
\;\in\;
\argmin_{\mu\in\Mm_+(\Zz)}
\left\{
\int_{\Zz} V(z)\,d\mu(z)+\Omega(\mu)
\right\}.
\end{equation}
Throughout this work, we assume that this minimizer is unique for all potentials
under consideration. Thus \(\mu_\Omega[V]\) is well-defined as a single measure.
For example, uniqueness is guaranteed when \(\Omega\) is strictly convex. We
study the inverse problem of recovering a continuous potential
\(V^\star\in\Cc(\Zz)\) from observations of the induced measure
\(\mu^\star = \mu_\Omega[V^\star]\).

\paragraph{Observations.}
We observe i.i.d.\ samples
\(z_1,\ldots,z_n \sim \overline{\mu}^{\star}\), where
\(\overline{\mu}^{\star}\eqdef \mu^\star/\mu^\star(\Zz)\) is the normalized
version of \(\mu^\star\). We denote the corresponding empirical measure by
\(\hat\mu^n\): it places mass \(m_{\mu^\star}/n\) on each sample \(z_i\), where
\(m_{\mu^\star}\eqdef \mu^\star(\Zz)\). We will say that $\hat\mu^n$ is an
$n$-sample empirical version of $\mu^\star$.

We assume that the total mass \(m_{\mu^\star}\) is known. This is
automatic, for example, when we work over probability measures. The extension to
estimated total masses should only introduce a straightforward bias term; for
clarity, we leave out these technicalities.

\paragraph{Parametrization.}
Let \(\phi_1,\ldots,\phi_S\in\Cc(\Zz)\) be a collection of basis functions. We
consider potentials of the form
\begin{equation}\label{eq:param}
    V_\theta(z)
    \eqdef
    \sum_{i=1}^S \theta_i \phi_i(z),
    \qquad
    \theta\in\RR^S.
\end{equation}
Our analysis focuses on this finite-dimensional parametrization, although
infinite-dimensional parametrizations are an interesting direction for future
work. The inverse problem can therefore be stated as follows: assuming that
\(V^\star=V_{\theta^\star}\), construct an estimator \(\hat\theta^n\) of
\(\theta^\star\) from empirical data \(\hat\mu^n\), and quantify the error
\(\norm{\hat\theta^n-\theta^\star}\) as a function of the number of samples
\(n\).


\subsection{Examples}

We now describe several settings in which the problem of estimating \(V^\star\)
from samples of \(\mu^\star\) arises naturally. These examples also fix the
notation used later for likelihood learning, inverse transport, unbalanced
matching, and inverse gradient flows.

\paragraph{Illustrative example: Maximum likelihood estimation.}
Suppose that \(\mu^\star = e^{-V_{\theta^\star}}\nu\), where \(\nu\) is a
reference measure, and \(V_\theta\) parametrizes the negative
log-density of a family of probability measures with respect to \(\nu\). Then the
forward problem corresponds to the density-estimation problem with
\(\Omega(\mu)=\iota_{\{\mu(\Zz)=1\}}(\mu)+\KL(\mu\mid\nu)\).
Throughout, for a set \(C\), \(\iota_C\) denotes the indicator function of \(C\),
equal to \(0\) on \(C\) and \(+\infty\) otherwise.

\paragraph{Illustrative example: Inverse optimal transport.}

Let \(\Zz=\Xx\times\Yy\), where \(\Xx\) and \(\Yy\) are complete separable metric
spaces. Given marginals \(\mu_1^\star\in\Pp(\Xx)\) and
\(\mu_2^\star\in\Pp(\Yy)\), define
\(\Omega:\Mm_+(\Xx\times\Yy)\to\overline{\RR}\) by
\begin{equation}\label{eq:iOT_Omega}
    \Omega(\mu)
    =
    \Omega_{\mu_1^\star,\mu_2^\star}(\mu)
    \eqdef
    \epsilon \KL(\mu\mid\mu_1^\star\otimes\mu_2^\star)
    +
    \iota_{\{\mu_1=\mu_1^\star\}}(\mu)
    +
    \iota_{\{\mu_2=\mu_2^\star\}}(\mu),
\end{equation}
where \(\mu_1\) and \(\mu_2\) denote the marginals of \(\mu\), and
\(\epsilon\ge 0\). Then~\eqref{eq:forward} is the entropy-regularized optimal
transport problem, and \(\mu^\star\) is the optimal coupling between
\(\mu_1^\star\) and \(\mu_2^\star\) with ground cost
\(V^\star\in\Cc(\Xx\times\Yy)\). In particular, when
\(V^\star(x,y)=\norm{x-y}^2\) and \(\epsilon=0\), this recovers the classical
quadratic optimal transport problem.

The inverse problem is to recover the cost \(V^\star\) from i.i.d.\ samples
\((x_i,y_i)\sim \mu^\star\), \(i=1,\ldots,n\), drawn from the optimal
coupling. In this case,
\(\hat\mu^n=n^{-1}\sum_{i=1}^n \delta_{(x_i,y_i)}\).

When \(\Xx\subseteq\RR^{d_1}\) and \(\Yy\subseteq\RR^{d_2}\), an important
parametric family is given by bilinear costs
\[
V_\theta(x,y)=-x^\top \theta y,
\qquad
\theta\in\RR^{d_1\times d_2}.
\]
In the notation of~\eqref{eq:param}, this corresponds to \(S=d_1d_2\), with
basis functions
\[
\phi_{i_1,i_2}(x,y)=x_{i_1}y_{i_2},
\qquad
i_1\in[d_1],\quad i_2\in[d_2].
\]

\paragraph{Running example: Inverse unbalanced optimal transport.}

More generally, one can relax the hard marginal constraints
in~\eqref{eq:iOT_Omega} and consider instead
\[
\Omega_{\nu_1^\star,\nu_2^\star}(\mu)
\eqdef
\epsilon \KL(\mu\mid\nu_1^\star\otimes\nu_2^\star)
+
D_{\varphi_1}(\mu_1\mid\nu_1^\star)
+
D_{\varphi_2}(\mu_2\mid\nu_2^\star),
\]
where \(D_{\varphi_1}\) and \(D_{\varphi_2}\) are
\(\varphi\)-divergences~\cite{sejourne2023unbalanced} associated with convex
functionals \(\varphi_1,\varphi_2\), such as total variation or
Kullback--Leibler divergence. In particular, $\varphi_i$ is  a convex, positive, lower-semi-continuous function $\varphi:(0,\infty)\to [0,\infty)$ that satisfies $\varphi(1)=0$, and its associated divergence is defined as 
\begin{align*}
D_\varphi(\alpha|\beta):=\int_\mathcal{X} \varphi\Big(\frac{d\alpha}{d\beta}(x)\Big)d\beta(x)+\varphi^\prime_\infty\int_\mathcal{X}d\alpha^\perp(x),
\end{align*} 	
where $\varphi^\prime_\infty=\lim_{x\to \infty}\varphi(x)/x$, and where, $\alpha^\perp$ is defined via the Radon-Nikodym-Lebesgue decomposition $\alpha=(d\alpha/d\beta)\beta+\alpha^\perp$ (see \cite{sejourne2019sinkhorn}). 

The observations are then empirical versions of
\(\mu^\star=\mu_{\Omega_{\nu_1^\star,\nu_2^\star}}(V^\star)\),
\(\nu_1^\star\), and \(\nu_2^\star\).
In the case of iOT, \(\nu_1^\star\) and \(\nu_2^\star\) are precisely the
marginals of \(\mu^\star\), and the divergences
\(D_{\varphi_1},D_{\varphi_2}\) are indicators matching the marginals exactly.
In the unbalanced setting, the marginals of \(\mu\) are instead penalized for
deviating from \(\nu_1^\star\) and \(\nu_2^\star\), but they are not constrained
to match them exactly.

The forward problem~\eqref{eq:forward} then becomes the entropy-regularized
unbalanced optimal transport problem. This formulation allows one to learn cost
functions from samples of matches between two measures whose total masses, or
marginals, need not agree exactly. The \textit{inverse} problem seeks to recover
\(V^\star\) from empirical versions of
\(\mu^\star,\nu_1^\star,\nu_2^\star\).

\paragraph{Running example: Inverse gradient flow.}

Suppose we observe snapshots
\(x_i^k \overset{\mathrm{i.i.d.}}{\sim} \mu^k\), \(i\in[n_k]\),
\(k\in[K]\), where \(\mu^k\in\Pp(\Zz)\). Assume that consecutive snapshots are linked by the
variational model
\[
\mu^{k+1}
=
\argmin_{\mu\in\Pp(\Zz)}
\left\{
\int_{\Zz} V^\star(z)\,d\mu(z)
+
\frac1\tau W_{2,\epsilon}^2(\mu,\mu^k)
\right\}.
\]
Here \(W_{2,\epsilon}^2\) denotes the entropy-regularized squared
Wasserstein-2 cost,
\[
W_{2,\epsilon}^2(\mu,\nu)
\eqdef
\inf_{\pi\in\Pp(\Zz\times\Zz)}
\left\{
\int_{\Zz\times\Zz} \norm{x-y}^2\,d\pi(x,y)
+
\epsilon \KL(\pi\mid\mu\otimes\nu)
\right\}.
\]
This is the Jordan--Kinderlehrer--Otto (JKO) scheme. Formally, when
\(\epsilon=0\), as \(\tau\to0\), one has \(\mu^k \approx \mu_{\tau k}\),
where \((\mu_t)_{t\ge0}\) solves the Wasserstein gradient flow equation
\[
\partial_t \mu_t
+
\operatorname{div}(\mu_t\nabla V^\star)
=
0.
\]

If we focus on a single pair of snapshots \((\mu^k,\mu^{k+1})\), then this fits
into the general framework~\eqref{eq:forward} by taking
\(\Omega(\mu)\eqdef \tau^{-1}W_{2,\epsilon}^2(\mu,\mu^k)\) and
\(\mu^\star=\mu^{k+1}\).
The inverse problem is to recover the potential function \(V^\star\) from
empirical versions of \(\mu^k\) and \(\mu^{k+1}\). In
this work, we focus on the regime \(\epsilon>0\), which allows us to exploit the
improved stability and sample-complexity properties of entropy regularization.
The unregularized case \(\epsilon=0\) is left for future work.

Unlike settings in which exact particle trajectories are observed, here we only
observe independent samples from each snapshot. In particular, the sample
\(x_i^k\) is not assumed to be paired with, or dependent on, the sample
\(x_i^{k+1}\).

\section{Optimality-based formulations of loss functions}\label{sec:losses}

We present a learning framework for recovering \(\theta^\star\) from the empirical
measure \(\hat\mu^n\). Although recovering \(V^\star\) from the forward
model~\eqref{eq:forward} is a nonlinear inverse problem, the variational
structure of the problem leads naturally to convex loss functions based on
optimality gaps.
We will also show that these gap-based losses improve upon classical
kernel-based losses not only through convexity, but also through better
conditioning.

\subsection{The Fenchel--Young loss}

The first loss family turns optimality itself into a regression signal: a
candidate potential is good when the observed measure nearly minimizes its
forward variational problem. Given \(V\in\Cc(\Zz)\) and
\(\mu\in\operatorname{dom}(\Omega)\), define the
Fenchel--Young gap associated with \(\Omega\) by
\[
\mathcal G_\Omega(V;\mu)
\eqdef
\langle V,\mu\rangle+\Omega(\mu)+\Omega^*(-V).
\]
Equivalently,
\[
\mathcal G_\Omega(V;\mu)
=
\langle V,\mu\rangle+\Omega(\mu)
-
\inf_{\nu\in\Mm_+(\Zz)}
\left\{
\langle V,\nu\rangle+\Omega(\nu)
\right\}.
\]
By the Fenchel--Young inequality,
\(\mathcal G_\Omega(V;\mu)\ge 0\). Moreover,
\(\mathcal G_\Omega(V;\mu)=0\) if and only if
\(\mu\in\argmin_{\nu\in\Mm_+(\Zz)}
\{\langle V,\nu\rangle+\Omega(\nu)\}\).

In the population inverse problem, the target measure is
\(\mu^\star=\mu_\Omega[V^\star]\). Thus it is natural to recover \(V^\star\)
by minimizing the gap \(V\mapsto \mathcal G_\Omega(V;\mu^\star)\).
Since the term \(\Omega(\mu^\star)\) does not depend on \(V\), we often work
with the constant-shifted Fenchel--Young loss
\begin{equation}\label{eq:FY_loss_population}
\mathcal L_{\Omega}^{\mathrm{FY}}(V;\mu^\star)
\eqdef
\langle V,\mu^\star\rangle
-
\inf_{\mu\in\Mm_+(\Zz)}
\left\{
\langle V,\mu\rangle+\Omega(\mu)
\right\}.
\end{equation}
This loss differs from the nonnegative gap \(\mathcal G_\Omega(V;\mu^\star)\)
only by the additive constant \(\Omega(\mu^\star)\). Therefore it has the same
minimizers in \(V\).

For a parametrized potential \(V_\theta\), we set
\(\mathcal J_{\Omega}^{\mathrm{FY}}(\theta;\mu^\star)\) to the composition
\(\mathcal L_{\Omega}^{\mathrm{FY}}(V_\theta;\mu^\star)\).
When the minimizer \(\mu_\Omega[V]\) is unique, the map
\(V\mapsto \mathcal L_{\Omega}^{\mathrm{FY}}(V;\mu^\star)\) is differentiable,
with
\[
\partial_V \mathcal L_{\Omega}^{\mathrm{FY}}(V;\mu^\star)
=
\mu^\star-\mu_\Omega[V].
\]
Consequently, if \(V_\theta=\sum_{i=1}^S\theta_i\phi_i\), then
\[
\nabla_\theta
\mathcal J_{\Omega}^{\mathrm{FY}}(\theta;\mu^\star)
=
\left(
\langle \phi_i,\mu^\star-\mu_\Omega[V_\theta]\rangle
\right)_{i=1}^S.
\]
Another attractive aspect of \(\Jj_\Omega^{\mathrm{FY}}\) is that it is convex
with respect to \(\theta\), since the infimum over affine functions is concave.

In practice, we do not observe \(\mu^\star\), but only the empirical measure
\(\hat\mu^n\). Moreover, the functional \(\Omega\) may itself be replaced by an
empirical approximation \(\hat\Omega_n\); see the examples described next. We
therefore define the empirical
Fenchel--Young loss
\begin{equation}\label{eq:FY_loss_empirical}
\widehat{\mathcal L}_{n}^{\mathrm{FY}}(V)
\eqdef
\mathcal L_{\hat\Omega_n}^{\mathrm{FY}}(V;\hat\mu^n)
=
\langle V,\hat\mu^n\rangle
-
\inf_{\mu\in\Mm_+(\Zz)}
\left\{
\langle V,\mu\rangle+\hat\Omega_n(\mu)
\right\}.
\end{equation}
The corresponding empirical parameter objective is
\(\widehat{\mathcal J}_{n}^{\mathrm{FY}}(\theta)
\eqdef \widehat{\mathcal L}_{n}^{\mathrm{FY}}(V_\theta)\). The estimator is
obtained by solving
\(\hat\theta^n\in\argmin_\theta\widehat{\mathcal J}_{n}^{\mathrm{FY}}(\theta)\).

\subsection{Examples of Fenchel--Young losses}\label{sec:examples_FY}

The abstract gap construction becomes concrete once the forward functional is
specified. The following examples show how the same definition recovers
likelihood learning, inverse transport, and inverse JKO objectives.

\subsubsection{Illustrative example: Maximum likelihood estimation.}

Consider the parametric density estimation setting with
\[
\Omega(\mu)
=
\KL(\mu\mid\nu)+\iota_{\{\mu(\Zz)=1\}}(\mu).
\]
Then
\[
\mathcal L_{\Omega}^{\mathrm{FY}}(V_\theta;\hat\mu^n)
=
\int_{\Zz} V_\theta(z)\,d\hat\mu^n(z)
-
\inf_{\mu\in\Pp(\Zz)}
\left\{
\int_{\Zz} V_\theta(z)\,d\mu(z)+\KL(\mu\mid\nu)
\right\}.
\]
The minimizer of the inner problem is the probability measure
\(\mu_\theta\) satisfying
\[
\frac{d\mu_\theta}{d\nu}
=
\frac{e^{-V_\theta}}{\int e^{-V_\theta}\,d\nu}.
\]
Therefore,
\[
\mathcal L_{\Omega}^{\mathrm{FY}}(V_\theta;\hat\mu^n)
=
-\int_{\Zz}\log\left(\frac{d\mu_\theta}{d\nu}\right)
\,d\hat\mu^n
\]
up to an additive constant independent of \(\theta\). Thus the Fenchel--Young
objective recovers the usual negative log-likelihood.

\subsubsection{Running example: Inverse unbalanced optimal transport.}

For inverse optimal transport, the goal is to recover a cost \(V^\star\) from
samples of an optimal coupling \(\mu^\star\). To define the corresponding
forward problem, let
\[
    \nu_1^\star\in\Mm_+(\Xx),
    \qquad
    \nu_2^\star\in\Mm_+(\Yy)
\]
be finite reference measures. For a cost \(V\in\Cc(\Xx\times\Yy)\), set
\(\Omega = \Omega_{\nu_1^\star,\nu_2^\star}^{\mathrm{UOT}}\), with
\[
    \Omega_{\nu_1^\star,\nu_2^\star}^{\mathrm{UOT}}(\mu)
    :=
    \epsilon \KL(\mu\mid\nu_1^\star\otimes\nu_2^\star)
    +
    D_{\varphi_1}(\mu_1\mid\nu_1^\star)
    +
    D_{\varphi_2}(\mu_2\mid\nu_2^\star),
    \qquad
    \mu\in\Mm_+(\Xx\times\Yy).
\]
The population matching measure is
\[
    \mu^\star
    =
    \mu_{\Omega_{\nu_1^\star,\nu_2^\star}^{\mathrm{UOT}}}
    [V_{\theta^\star}].
\]

The balanced and unbalanced settings differ in how the reference measures are
observed. In the balanced inverse OT case, the reference marginals are the
marginals of the observed coupling. Thus one observes paired samples
\((x_i,y_i)_{i=1}^n\overset{\mathrm{i.i.d.}}{\sim}\mu^\star\),
where, according to the convention above, this means sampling from the normalized
measure $\bar\mu^\star$. The empirical coupling is $\widehat\mu_n$, and the
empirical marginals are obtained by projection:
\(\widehat\nu_{1,n}:=(\mathrm{pr}_{\Xx})_\#\widehat\mu_n\) and
\(\widehat\nu_{2,n}:=(\mathrm{pr}_{\Yy})_\#\widehat\mu_n\).

In the genuinely unbalanced case, the matching measure and the reference
measures are distinct finite measures. We therefore assume independent samples
from their normalized versions:
\((x_i,y_i)_{i=1}^{n_\mu}\overset{\mathrm{i.i.d.}}{\sim}\bar\mu^\star\),
\(\pa{a_i}_{i=1}^{n_1}\overset{\mathrm{i.i.d.}}{\sim}\bar\nu_1^\star\), and
\(\pa{b_j}_{j=1}^{n_2}\overset{\mathrm{i.i.d.}}{\sim}\bar\nu_2^\star\).
The corresponding mass-rescaled empirical measures are denoted
\(\widehat\mu_{n_\mu}\), \(\widehat\nu_{1,n_1}\), and
\(\widehat\nu_{2,n_2}\).

The empirical UOT functional is
\[
    \widehat\Omega_n(\mu)
    :=
    \epsilon \KL(\mu\mid \widehat\nu_{1,n_1}\otimes \widehat\nu_{2,n_2})
    +
    D_{\varphi_1}(\mu_1\mid \widehat\nu_{1,n_1})
    +
    D_{\varphi_2}(\mu_2\mid \widehat\nu_{2,n_2}).
\]
In the balanced case, where
\(D_{\varphi_i}(\mu_i\mid\nu_i)=\iota_{\{\mu_i=\nu_i\}}\), we take
\(\widehat\nu_{1,n_1}\) and \(\widehat\nu_{2,n_2}\) to be precisely the
marginals of \(\widehat\mu_{n_\mu}\).

The empirical inverse UOT objective is
\begin{align*}
\hat \Jj_n(\theta) = \widehat{\mathcal L}_{n}^{\mathrm{FY}}(V_\theta)
&=
\int_{\Xx\times\Yy} V_\theta(x,y)\,d\widehat\mu_{n_\mu}(x,y)
\\
&\quad
-
\inf_{\mu\in\Mm_+(\Xx\times\Yy)}
\Bigg\{
\int_{\Xx\times\Yy} V_\theta(x,y)\,d\mu(x,y)
+
\epsilon\KL(\mu\mid \widehat\nu_{1,n_1}\otimes\widehat\nu_{2,n_2})
\\
&\hspace{4cm}
+
D_{\varphi_1}(\mu_1\mid \widehat\nu_{1,n_1})
+
D_{\varphi_2}(\mu_2\mid \widehat\nu_{2,n_2})
\Bigg\}.
\end{align*}

\subsubsection{Running example: inverse gradient flow}\label{sec:fy_iGF}

We next consider the inverse problem associated with a single step of an
entropy-regularized JKO scheme. Let \(\epsilon>0\) be an entropic regularization
parameter and suppose that
\begin{equation*}
\mu^{k+1}
=
\argmin_{\mu\in\Pp(\Zz)}
\left\{
\int_{\Zz} V(z)\,d\mu(z)
+
\Omega(\mu)
\right\},
\qquad \Omega\eqdef
\Omega_{\mu^k}
\eqdef
\frac1\tau W_{2,\epsilon}^2(\cdot,\mu^k).
\tag{JKO}\label{eq:JKO}
\end{equation*}
Here \(\mu^k\) is the previous snapshot and \(\mu^{k+1}\) is the next snapshot.

In practice, we observe empirical snapshots
\(\hat\mu_k=n_k^{-1}\sum_i\delta_{x_i^k}\) and the analogous
\(\hat\mu_{k+1}\) built from \(x_i^{k+1}\).
The empirical inverse JKO Fenchel--Young loss is obtained by replacing
\((\mu^k,\mu^{k+1})\) in the population loss by
\((\hat\mu^k,\hat\mu^{k+1})\):
\begin{align}\label{eq:iJKO_loss}
\widehat{\mathcal L}_{n}^{\mathrm{FY}}
(V;\hat \mu^k,\hat \mu^{k+1})
&\eqdef
\langle V,\hat \mu^{k+1}\rangle
-
\inf_{\mu\in\Pp(\Zz)}
\left\{
\langle V,\mu\rangle
+
\frac1\tau W_{2,\epsilon}^2(\mu,\hat \mu^k)
\right\}.
\end{align}

The curvature and stability of the Fenchel--Young loss are inherited entirely
from the forward functional \(\Omega\). For the natural \(\Omega\) arising in
the JKO problem here, the nonlinear part of the Fenchel--Young loss need not
exhibit strong curvature.

The following simple example shows that this nonlinear term can be a hard
minimum and therefore piecewise linear.
Let
\begin{equation}\label{eq:example_2point}
    \Zz=\{0,1\},
\qquad
\mu^k=\delta_0,
\qquad
V(0)=V_0,\quad V(1)=V_1.
\end{equation}
For \(\mu=p\delta_0+(1-p)\delta_1\), with \(p\in[0,1]\), the coupling constraint in
\(W_{2,\epsilon}^2(\mu,\delta_0)\) forces
\(\pi=\mu\otimes\delta_0\). Hence
\(W_{2,\epsilon}^2(\mu,\delta_0)=1-p\) and
\(\Omega_k(\mu)=(1-p)/\tau\).
Thus the nonlinear term is
\(F_0(V)=\inf_{p\in[0,1]}\{pV_0+(1-p)V_1+(1-p)/\tau\}
=\min\{V_0,\;V_1+1/\tau\}\).
Consequently, the loss is convex but piecewise linear, and therefore has no
local strong convexity away from the kink. In
Section~\ref{sec:sharpened-fy-losses}, we address this issue by introducing
the idea of sharpening.

\subsection{The sharpened Fenchel--Young loss}\label{sec:sharpened-fy-losses}

Fenchel--Young losses provide convex objectives, but their curvature and
stability properties are inherited entirely from the forward functional
\(\Omega\). In particular, the observation \(\hat\mu\) enters the loss only
through the linear term \(\langle V,\hat\mu\rangle\). This may be insufficient
for robustness in noisy settings, especially in the inverse gradient flow problem
studied below.

To address this limitation, we modify the forward variational problem by adding
a data-dependent discrepancy term. This changes the geometry of the optimality
conditions while preserving the gap-loss structure and the calibration of the
original inverse problem.

Let \(D:\Mm_+(\Zz)\times\Mm_+(\Zz)\to[0,+\infty]\) be a discrepancy satisfying
\(D(\mu\mid\nu)\ge 0\) for all \(\mu,\nu\in\Mm_+(\Zz)\), and
\(D(\nu\mid\nu)=0\). Given a reference measure \(\nu\), define the sharpened
functional \(\Omega^\sharp_{\nu}(\mu)\eqdef \Omega(\mu)+D(\mu\mid\nu)\).

\begin{defn}[Sharpened Fenchel--Young loss]\label{defn:gen_fy}
For \(V\in\Cc(\Zz)\) and \(\nu\in\operatorname{dom}(\Omega)\), the sharpened
Fenchel--Young gap is
\[
\mathcal G_{\Omega^\sharp_\nu}(V;\nu)
\eqdef
\langle V,\nu\rangle
+
\Omega^\sharp_{\nu}(\nu)
-
\inf_{\mu\in\Mm_+(\Zz)}
\left\{
\langle V,\mu\rangle+ \Omega^\sharp_{\nu}(\mu)
\right\}.
\]
We define
\[
\mathcal L_{D,\Omega}^{\sharp}(V;\nu) \eqdef
\mathcal L_{\Omega^\sharp_\nu}^{\mathrm{FY}}(V;\nu)
\eqdef
\mathcal G_{\Omega^\sharp_\nu}(V;\nu).
\]
We use $\mathcal L_{D,\Omega}^{\sharp}$ when emphasizing the act of sharpening,
and \(\mathcal L_{\Omega^\sharp_\nu}^{\mathrm{FY}}\) when treating the
sharpened problem as an ordinary Fenchel--Young loss for a modified functional.
\end{defn}

The discrepancy term allows the data to influence the nonlinear part of the
loss. As shown in subsequent sections, this can yield improved local curvature
and stability properties, which are essential for statistical recovery. We
summarize the basic properties of the sharpened loss next.

\begin{prop}[Basic properties]\label{prop:key_properties}
Fix \(\hat\mu\in\operatorname{dom}(\Omega)\), and suppose that
\(D(\mu\mid\hat\mu)\ge 0\) for all \(\mu\), with
\(D(\hat\mu\mid\hat\mu)=0\). Then
\[
V\mapsto \mathcal L_{D,\Omega}^{\sharp}(V;\hat\mu)
\]
satisfies:
\begin{enumerate}
    \item Convexity in \(V\).

    \item Non-negativity:
    \[
     \mathcal L_{D,\Omega}^{\sharp}(V;\hat\mu)\ge 0.
    \]

    \item Its zeros are characterized by the sharpened optimality condition:
    \[
     \mathcal L_{D,\Omega}^{\sharp}(V;\hat\mu)=0
    \quad\Longleftrightarrow\quad
    \hat\mu\in
    \argmin_{\mu}
    \left\{
    \langle V,\mu\rangle+\Omega(\mu)+D(\mu\mid\hat\mu)
    \right\}.
    \]

    \item Calibration with respect to the original forward problem: if
    \[
    \hat\mu\in
    \argmin_{\mu}
    \left\{
    \langle V,\mu\rangle+\Omega(\mu)
    \right\},
    \]
    then
    \[
    \mathcal L_{D,\Omega}^{\sharp}(V;\hat\mu)=0.
    \]
    Hence sharpening preserves all exact solutions of the original inverse
    problem.

    \item If \(\Omega^\sharp_{\hat\mu}\) is strictly convex, then the loss is
    differentiable in \(V\), with
    \[
    \partial_V \mathcal L_{\Omega^\sharp_{\hat\mu}}^{\mathrm{FY}}(V;\hat\mu)
    =
    \hat\mu-\mu_{\Omega,D}[V;\hat\mu],
    \]
    where
    \[
    \mu_{\Omega,D}[V;\hat\mu]
    \eqdef
    \argmin_\mu
    \left\{
    \langle V,\mu\rangle+\Omega(\mu)+D(\mu\mid\hat\mu)
    \right\}.
    \]

    \item Monotonicity in the sharpening term: if \(0\le D'\le D\), then
    \[
    \mathcal L_{D',\Omega}^{\sharp}(V;\hat\mu)
    \ge
    \mathcal L_{D,\Omega}^{\sharp}(V;\hat\mu).
    \]
\end{enumerate}
\end{prop}

\begin{rem}[Related losses]
The sharpened Fenchel--Young loss directly generalizes several existing notions.
\begin{itemize}
    \item[(i)]
    When \(D\equiv 0\), the sharpened loss reduces to the classical
    Fenchel--Young loss associated with \(\Omega\). In this case, evaluating the
    loss requires computing \(\Omega^*(-V)\), equivalently solving the forward
    problem~\eqref{eq:forward}.

    \item[(ii)]
    Fitzpatrick losses arise as a special case of sharpened Fenchel--Young
    losses when \(D\) is chosen as the Bregman divergence associated with
    \(\Omega\). In this case, the sharpened loss coincides with the Fitzpatrick
    function of the associated monotone operator. Existing comparisons between
    Fenchel--Young and Fitzpatrick losses can be interpreted through the
    monotonicity property in Proposition~\ref{prop:key_properties}.
\end{itemize}
\end{rem}

In practice, we use an empirical version of the sharpened loss, where sampling
is incorporated both through \(\hat\Omega_n\) and through the sharpening term.
As before, we drop additive constants independent of \(V\). Given data
\(\hat\mu^n\) and an
empirical functional \(\hat\Omega_n\), the empirical loss is
\begin{equation}\label{eq:sharpened_FY}
\widehat{\mathcal L}_{n}^\sharp(V)
\eqdef
\langle V,\hat\mu^n\rangle
-
\inf_{\mu\in\Mm_+(\Zz)}
\left\{
\langle V,\mu\rangle
+
\hat\Omega_n(\mu)
+
D(\mu\mid\hat\mu^n)
\right\}.
\end{equation}
The corresponding empirical parameter objective is then
$\widehat{\mathcal J}_{n}(\theta)
\eqdef
\widehat{\mathcal L}_{n}^\sharp(V_\theta)$.

\subsection{Examples of sharpened Fenchel--Young losses}

We now illustrate the effect of sharpening in two settings. The first example
shows explicitly how sharpening improves the curvature of a likelihood-based
objective. The second example explains why sharpening is useful for inverse
gradient-flow problems, where the unsharpened objective may have weak curvature.

\subsubsection{Illustrative example: maximum-likelihood estimation}
\label{sec:precision_sharpening_example}

We first consider a simple maximum-likelihood estimation problem. The forward
model corresponds to entropic density estimation, which fits into our framework
by taking
\[
\Omega(\mu)=\KL(\mu\mid dx)+\iota_{\{\mu(\Zz)=1\}}(\mu).
\]
Thus, for a potential \(V\), the forward map is
\[
\mu_\Omega[V]
=
\argmin_{\mu\in\Pp(\Zz)}
\left\{
\langle V,\mu\rangle+\KL(\mu\mid dx)
\right\},
\]
so that \(\mu_\Omega[V]\) has density proportional to \(e^{-V}\). In this case,
the ordinary Fenchel--Young loss coincides, up to an additive constant, with the
negative log-likelihood.

\paragraph{Precision matrix estimation.}

Let the population observation be
\[
\mu^\star=\mathcal N(0,\Sigma^\star),
\qquad
\Sigma^\star\succ0,
\]
and parameterize the model by a precision matrix \(K\succ0\) via
\[
V_K(x)=\frac12 x^\top Kx.
\]
Then \(\mu_\Omega[V_K]=\mathcal N(0,K^{-1})\).
We now sharpen the model using a KL discrepancy to the observed population
distribution. Define
\[
D_r(\mu\mid\mu^\star)
\eqdef
r\,\KL(\mu\mid\mu^\star),
\qquad r>0.
\]
The sharpened functional is therefore
\[
\Omega^\sharp_{\mu^\star}(\mu)
=
\KL(\mu\mid dx)+\iota_{\{\mu(\Zz)=1\}}(\mu)
+
r\KL(\mu\mid\mu^\star).
\]
To evaluate the sharpened loss
\(\mathcal J_r(K)\eqdef
\mathcal L_{D_r,\Omega}^{\sharp}(V_K;\mu^\star)\), write
\(p^\star=d\mu^\star/dx\). The inner variational problem in
\(\mathcal J_r\) satisfies
\[
\inf_{\mu\in\Pp(\Zz)}
\left\{
\langle V_K,\mu\rangle
+
\KL(\mu\mid dx)
+
r\KL(\mu\mid\mu^\star)
\right\}
=
-(1+r)\log
\int
\exp\!\left(-\frac{V_K(x)}{1+r}\right)
p^\star(x)^{r/(1+r)}\,dx.
\]
Substituting \(\mu^\star=\mathcal N(0,\Sigma^\star)\) gives
\[
(1+r)\log
\int
\exp\!\left(-\frac{V_K(x)}{1+r}\right)
p^\star(x)^{r/(1+r)}\,dx
=
-\frac{1+r}{2}
\log\det\!\left(K+r(\Sigma^\star)^{-1}\right)
+
\mathrm{const}.
\]
Consequently,
\begin{equation}\label{eq:precision_sharpened_loss}
\mathcal J_r(K)
=
\frac12
\left(
\tr(K\Sigma^\star)
-
(1+r)\log\det\!\left(K\Sigma^\star+rI\right)
\right)
+
\mathrm{const}.
\end{equation}

Introduce the whitened precision matrix
\(A=(\Sigma^\star)^{1/2}K(\Sigma^\star)^{1/2}\).
Up to an additive constant, the objective becomes
\[
\mathcal J_r(A)
=
\frac12
\left(
\tr A
-
(1+r)\log\det(A+rI)
\right).
\]
Thus, along an eigendirection with eigenvalue \(\lambda>0\), the scalar objective
is \(j_r(\lambda)=\frac12(\lambda-(1+r)\log(\lambda+r))\), and its curvature is
\(j_r''(\lambda)=(1+r)/(2(\lambda+r)^2)\).
Hence sharpening replaces the singular log-determinant curvature
\(\lambda^{-2}\) when $r=0$ by the regularized curvature
\((1+r)(\lambda+r)^{-2}\).
Over a spectral range \(\lambda\in[m,M]\), the curvature condition number is
therefore improved from
\(\left(\frac{M}{m}\right)^2\)
to \(\left(\frac{M+r}{m+r}\right)^2\).
Thus KL sharpening shifts the log-determinant barrier away from the boundary and
improves spectral conditioning of the precision-estimation objective.

\subsubsection{Running example: inverse gradient flow}\label{sec:sharpened_iGF}

We return to the inverse JKO problem discussed in~\eqref{eq:JKO}.
To improve stability and regularity of the inverse objective, we introduce a
discrepancy that penalizes deviations from the observed next snapshot:
\[
D_r(\mu\mid\mu^{k+1})
\eqdef
r\,\KL(\mu\mid\mu^{k+1}),
\qquad r>0.
\]
The sharpened forward functional is then
\[
\Omega^\sharp_{\mu^{k+1}}(\mu)
\eqdef
\Omega_k(\mu)
+
D_r(\mu\mid\mu^{k+1})
=
\frac1\tau W_{2,\epsilon}^2(\mu,\mu^k)
+
r\,\KL(\mu\mid\mu^{k+1}).
\]
Dropping additive constants independent of \(V\), the associated sharpened
Fenchel--Young loss is
\begin{align}\label{eq:sharpened_iJKO_loss}
\mathcal L_{\mathrm{iJKO},r}^{\sharp}
(V;\mu^k,\mu^{k+1})
&\eqdef
\langle V,\mu^{k+1}\rangle
\\
&\quad
-
\inf_{\mu\in\Pp(\Zz)}
\left\{
\langle V,\mu\rangle
+
\frac1\tau W_{2,\epsilon}^2(\mu,\mu^k)
+
r\,\KL(\mu\mid\mu^{k+1})
\right\}.
\end{align}
By the calibration property of the sharpened Fenchel--Young loss, if
\(\mu^{k+1}\) solves the original JKO step associated with \(V\), then
\[
\mathcal L_{\mathrm{iJKO},r}^{\sharp}
(V;\mu^k,\mu^{k+1})
=0.
\]
Thus the KL sharpening term preserves exact solutions of the original inverse
problem, while modifying the local geometry of the loss.

In practice, given empirical snapshots
$\hat\mu_k
=
\frac1{n_k}\sum_{i=1}^{n_k}\delta_{x_i^k}$ and $
\hat\mu_{k+1}
=
\frac1{n_{k+1}}\sum_{i=1}^{n_{k+1}}\delta_{x_i^{k+1}}$,
the empirical sharpened inverse JKO loss is obtained by replacing
\((\mu^k,\mu^{k+1})\) in~\eqref{eq:sharpened_iJKO_loss} by
\((\hat\mu_k,\hat\mu_{k+1})\):
\[
\widehat{\mathcal L}_{\mathrm{iJKO},r}^{\sharp}(V)
\eqdef
\mathcal L_{\mathrm{iJKO},r}^{\sharp}
(V;\hat\mu_k,\hat\mu_{k+1}).
\]
This empirical notation will be used again in the stability theorem and in the
numerical implementation.

\paragraph{Why sharpening is useful.}

The KL sharpening term regularizes the inner variational problem by anchoring
the optimizer to the observed next snapshot. Returning to the two-point example
from~\eqref{eq:example_2point}, we saw that,
without sharpening, the nonlinear part of the Fenchel--Young loss is a hard
minimum and is therefore piecewise linear. We now show that KL sharpening turns
it into a smooth soft minimum with positive curvature in the identifiable
direction.

Let
\[
\mu^{k+1}=q\delta_0+(1-q)\delta_1,
\qquad q\in(0,1),
\]
and let \(r>0\). The nonlinear part of the sharpened Fenchel--Young loss is
\[
F_r(V)
\eqdef
\inf_{p\in[0,1]}
\left\{
pV_0+(1-p)V_1+\frac{1-p}{\tau}
+
r\,\KL
\bigl(
p\delta_0+(1-p)\delta_1
\,\big|\,
\mu^{k+1}
\bigr)
\right\}.
\]
Then
\[
F_r(V)
=
-r\log
\left(
q e^{-V_0/r}
+
(1-q)e^{-(V_1+1/\tau)/r}
\right).
\]
Moreover, with
\[
\rho_r(V)
\eqdef
\frac{q e^{-V_0/r}}
{
q e^{-V_0/r}
+
(1-q)e^{-(V_1+1/\tau)/r}
},
\]
one has
\[
-\nabla^2 F_r(V)
=
\frac{\rho_r(V)(1-\rho_r(V))}{r}
\begin{pmatrix}
1 & -1\\
-1 & 1
\end{pmatrix}.
\]
Hence the sharpened loss
\(\mathcal L_{\mathrm{iJKO},r}^{\sharp}(V;\mu^k,\mu^{k+1})
=\langle V,\mu^{k+1}\rangle-F_r(V)\) is smooth in \(V\). The sharpened loss is
not strongly convex on all of
\(\RR^2\), because it is
invariant under adding constants to \(V\). However, it is locally strongly convex
on the quotient space \(\RR^2/\operatorname{span}\{(1,1)\}\). Equivalently,
after fixing a gauge, for instance by imposing \(V_0+V_1=0\), the sharpened
loss is locally strongly convex.

\subsection{Relation to quadratic losses}

Quadratic matching losses are often attractive computationally, but they can
hide the variational structure of the inverse problem. This subsection explains
when such losses should be viewed as preconditioned or limiting versions of the
gap objectives introduced above.

\subsubsection{MMD}
A classical approach is to compare the observed empirical measure
\(\hat\mu^n\) with the forward prediction \(\mu_{\hat\Omega_n}[V_\theta]\) by
minimizing a kernel discrepancy:
\begin{equation}\label{eq:mmd}
\min_\theta
\left\|
\mu_{\hat\Omega_n}[V_\theta]-\hat\mu^n
\right\|_\kappa^2 \qwhereq \mu_{\hat\Omega_n}[V_\theta]
\in
\argmin_{\mu\in\Mm_+(\Zz)}
\left\{
\int V_\theta(z)\,d\mu(z)+\hat\Omega_n(\mu)
\right\},
\end{equation}
and \(\|\rho\|_\kappa^2
\eqdef \iint \kappa(z,z')\,d\rho(z)\,d\rho(z')\) is the squared MMD norm
associated with a positive definite kernel
\(\kappa\)~\cite{smola2007hilbert}.
Unlike the Fenchel--Young objective, this formulation is generally 
nonconvex in \(\theta\).

Nevertheless, the MMD objective is closely related to the gradient of the
Fenchel--Young loss. Suppose that
\(V_\theta=\Phi\theta\eqdef \sum_{k\in\NN}\theta_k\phi_k\), and that the
kernel admits the feature expansion
\(\kappa(z,z')=\sum_{j\in\NN}\psi_j(z)\psi_j(z')\).
Define the MMD residual coefficients
\[
w_j(\theta)
\eqdef
\left\langle
\psi_j,
\mu_{\hat\Omega_n}[\Phi\theta]-\hat\mu^n
\right\rangle,
\qquad
\mathbf w(\theta)\eqdef (w_j(\theta))_{j\in\NN}.
\]
Then
\[
\mathcal J_{\mathrm{MMD}}(\theta)
\eqdef
\left\|
\mu_{\hat\Omega_n}[\Phi\theta]-\hat\mu^n
\right\|_\kappa^2
=
\|\mathbf w(\theta)\|_{\ell_2}^2.
\]

On the other hand, the empirical Fenchel--Young objective
\(\widehat{\mathcal J}_{n}^{\mathrm{FY}}(\theta)
\eqdef \widehat{\mathcal L}_{n}^{\mathrm{FY}}(\Phi\theta)\) satisfies
\[
\nabla_\theta \widehat{\mathcal J}_{n}^{\mathrm{FY}}(\theta)
=
\left(
\left\langle
\phi_k,
\hat\mu^n-\mu_{\hat\Omega_n}[\Phi\theta]
\right\rangle
\right)_{k\in\NN}.
\]
Assume that each \(\phi_k\) admits the expansion
$\phi_k
=
\sum_{j\in\NN} c_{k,j}\psi_j$ and
let \(C_{\Phi,\Psi} = (c_{k,j})_{k,j\in\NN}\) be the change of basis operator. Then,
\[
\nabla_\theta \widehat{\mathcal J}_{n}^{\mathrm{FY}}(\theta)
=
- C_{\Phi,\Psi}\mathbf w(\theta).
\]
Consequently, whenever \(C_{\Phi,\Psi}\) is invertible,
\[
\mathcal J_{\mathrm{MMD}}(\theta)
=
\left\|
C_{\Phi,\Psi}^{-1}
\nabla_\theta \widehat{\mathcal J}_{n}^{\mathrm{FY}}(\theta)
\right\|_{\ell_2}^2,
\]
and, wherever the Fenchel--Young objective is twice differentiable,
\[
\nabla_\theta \mathcal J_{\mathrm{MMD}}(\theta)
=
2\,\nabla_\theta^2 \widehat{\mathcal J}_{n}^{\mathrm{FY}}(\theta)\,
C_{\Phi,\Psi}^{-\top}C_{\Phi,\Psi}^{-1}
\nabla_\theta \widehat{\mathcal J}_{n}^{\mathrm{FY}}(\theta).
\]

This identity shows that the MMD objective is a squared, preconditioned gradient
norm of the Fenchel--Young objective. Thus any poor conditioning in the curvature
of \(\widehat{\mathcal J}_{n}^{\mathrm{FY}}\) can be amplified by the quadratic
least-squares formulation. In contrast, directly minimizing the
Fenchel--Young objective exploits the convexity and local curvature of the
underlying optimality gap. Moreover, studying the structure of
\(\widehat{\mathcal J}_{n}^{\mathrm{FY}}\), for instance establishing local
strong convexity, can directly inform the behavior of the MMD loss.

\subsubsection{Connection with the quadratic iJKO\texorpdfstring{\(^\star\)}{*} loss}
\label{sec:limiting_jko_loss}

Terpin et al.~\cite{terpin2024learning} proposed a quadratic inverse-JKO loss
derived from the first-order optimality condition of the JKO step. In the
unregularized smooth setting, if \(\mu^k\) and \(\mu^{k+1}\) are connected by an
optimal transport map \(T:\Zz\to\Zz\) pushing \(\mu^{k+1}\) to \(\mu^k\), then
the JKO optimality condition gives
\(\nabla V(y)+\tau^{-1}(y-T(y))=0\) \(\mu^{k+1}\)-a.e.
This motivates the quadratic loss
\begin{equation}
\label{eq:ijko-star-loss}
    \mathcal J_{\mathrm{iJKO}^\star}(\theta)
    \eqdef
    \int
    \left\|
        \nabla V_\theta(y)
        +
        \frac1\tau (y-T(y))
    \right\|^2
    d\mu^{k+1}(y).
\end{equation}
Equivalently, if \(f^\ast_{k+1,k}\) is a Kantorovich potential from
\(\mu^{k+1}\) to \(\mu^k\), then \(y-T(y)=\nabla f^\ast_{k+1,k}(y)\), and
\[
    \mathcal J_{\mathrm{iJKO}^\star}(\theta)
    =
    \int
    \left\|
        \nabla V_\theta
        +
        \tau^{-1}\nabla f^\ast_{k+1,k}
    \right\|^2
    d\mu^{k+1}.
\]
Thus iJKO\(^\star\) first estimates an OT map or Kantorovich potential from the
two snapshots and then fits \(V_\theta\) by least squares. By contrast, the
Fenchel--Young approach optimizes the variational gap directly from the two
snapshots and keeps the convex structure of the JKO objective.


On the other hand,
the high-sharpening limit can be interpreted as a local quadratic
approximation to the Fenchel--Young geometry. In sharpened iJKO, the large KL weight
forces the inner optimizer to remain close to the observed next snapshot
\(\mu^{k+1}\). Localized around \(\mu^{k+1}\), the nonlinear
part of the JKO objective can be replaced, to first order, by its first
variation at \(\mu^{k+1}\). This gives the effective linear statistic
\(h_\theta=V_\theta+\tau^{-1}f^\ast_{k+1,k}\), up to additive constants. As we
show below, the sharpened Fenchel--Young loss
converges to a quadratic residual involving the Kantorovich potential.

This perspective clarifies both the usefulness and the limitations of
quadratic inverse losses. They can be viewed as computationally convenient
first-order surrogates for the full variational objectives, and in regimes
where the optimizer is strongly localized they capture the leading-order
statistical signal. At the same time, they may discard global information
contained in the original Fenchel--Young geometry, including nonlinear
transport effects that can be
important for identifiability and consistency away from the limiting regime.

\begin{prop}[High-sharpening limit]
\label{prop:r_infty_ijko}
Let
\(F_r(\theta)\eqdef \mathcal L_{\mathrm{iJKO},r}^{\sharp}
(V_\theta;\mu^k,\mu^{k+1})\), and let \(f^\ast_{k+1,k}\) be the entropic
Kantorovich potential associated with the transport problem from
\(\mu^{k+1}\) to \(\mu^k\). Set
\(h_\theta\eqdef V_\theta+\tau^{-1}f^\ast_{k+1,k}\).
Since the variance is invariant under additive constants, the normalization of
\(f^\ast_{k+1,k}\) is immaterial. Then
\begin{equation}
\label{eq:limit_fy2}
\lim_{r\to\infty}
r\left(
F_r(\theta)
+
\frac1\tau
W_{2,\epsilon}^2
(\mu^{k+1},\mu^k)
\right)
=
\frac12
\operatorname{Var}_{\mu^{k+1}}
\left[h_\theta\right].
\end{equation}
Moreover, if
\[
    \theta_r^\lambda
    \in
    \argmin_\theta
    \{F_r(\theta)+\lambda R(\theta)\}
    \qquad\text{and}\qquad
    \lambda=\lambda_0/r,
\]
then every cluster point of \(\theta_r^{\lambda_0/r}\) minimizes
\[
    \frac12
    \operatorname{Var}_{\mu^{k+1}}\left[h_\theta\right]
    +
    \lambda_0 R(\theta).
\]
\end{prop}

\begin{proof}
Define
\[
G(\theta,\mu)
\eqdef
\langle V_\theta,\mu\rangle
+
\frac1\tau
W_{2,\epsilon}^2(\mu,\mu^k),
\]
and set
\[
\widetilde F_r(\theta)
\eqdef
F_r(\theta)
+
\frac1\tau
W_{2,\epsilon}^2(\mu^{k+1},\mu^k).
\]
Then
\[
\widetilde F_r(\theta)
=
G(\theta,\mu^{k+1})
-
\inf_{\mu\in\Pp(\Zz)}
\left\{
G(\theta,\mu)
+
r\,\KL(\mu\mid\mu^{k+1})
\right\}.
\]
Let
\[
    \mu_r^\theta
    \eqdef
    \mu_{\Omega_k,D_r}[V_\theta;\mu^{k+1}]
\]
denote the minimizer of the inner problem, where
\(\Omega_k(\mu)=\tau^{-1}W_{2,\epsilon}^2(\mu,\mu^k)\) and
\(D_r(\mu\mid\mu^{k+1})=r\,\KL(\mu\mid\mu^{k+1})\). As
\(r\to\infty\), the KL penalty forces \(\mu_r^\theta\to\mu^{k+1}\), and
\(\widetilde F_r(\theta)\to0\). Set \(t=1/r\) and
\[
    g_\theta(t)\eqdef \widetilde F_{1/t}(\theta),
    \qquad
    \mu_t^\theta\eqdef \mu_{1/t}^\theta.
\]
Then \(g_\theta(0)=0\) and
\[
    \lim_{r\to\infty} r\widetilde F_r(\theta)
    =
    \lim_{t\downarrow0}\frac{g_\theta(t)-g_\theta(0)}{t}
    =
    g_\theta'(0),
\]
provided the right derivative exists. The envelope theorem gives, for
\(t>0\),
\[
    g_\theta'(t)
    =
    \frac{1}{t^2}
    \KL(\mu_t^\theta\mid\mu^{k+1}).
\]
Indeed, the only explicit dependence on \(t\) in the inner minimization is the
coefficient \(1/t\) in front of the KL term, and differentiating
\(-\inf_\mu\{G(\theta,\mu)+t^{-1}\KL(\mu\mid\mu^{k+1})\}\) yields the displayed
identity.

It remains to compute the limit of \(t^{-2}\KL(\mu_t^\theta\mid\mu^{k+1})\).
By differentiability of entropic transport in the first marginal,
\[
    \frac{\delta G}{\delta\mu}(\theta,\mu)
    =
    V_\theta+\tau^{-1}f^\ast(\mu,\mu^k),
\]
where \(f^\ast(\mu,\mu^k)\) denotes an entropic Kantorovich potential from
\(\mu\) to \(\mu^k\). The first-order condition for \(\mu_t^\theta\) gives the
exact exponential tilt
\[
    \frac{d\mu_t^\theta}{d\mu^{k+1}}
    =
    \frac{
        \exp\!\left(
            -t\left[
                V_\theta+\tau^{-1}f^\ast(\mu_t^\theta,\mu^k)
            \right]
        \right)
    }{
        \int
        \exp\!\left(
            -t\left[
                V_\theta+\tau^{-1}f^\ast(\mu_t^\theta,\mu^k)
            \right]
        \right)
        d\mu^{k+1}
    }.
\]
Since \(\mu_t^\theta\to\mu^{k+1}\) and the entropic Kantorovich potential is
continuous with respect to the first marginal, the tilting function
\[
    V_\theta+\tau^{-1}f^\ast(\mu_t^\theta,\mu^k)
\]
converges in \(L^2(\mu^{k+1})\), modulo constants, to \(h_\theta\). Applying
Lemma~\ref{lem:kl_exp} with this convergent family of tilts therefore yields
\[
    \lim_{t\downarrow0}
    \frac{1}{t^2}
    \KL(\mu_t^\theta\mid\mu^{k+1})
    =
    \frac12
    \operatorname{Var}_{\mu^{k+1}}[h_\theta].
\]
Hence \(g_\theta'(0)=\frac12
\operatorname{Var}_{\mu^{k+1}}[h_\theta]\), which is
\eqref{eq:limit_fy2}. The statement about minimizers follows by the
corresponding \(\Gamma\)-convergence of
\(r\widetilde F_r+\lambda_0 R\).
\end{proof}

\begin{lem}[KL expansion under exponential tilting]
\label{lem:kl_exp}
Let \(q\in\Pp(\Zz)\) and let \(g\in L^2(q)\). Define \(p_t\ll q\) by
\[
    \frac{dp_t}{dq}
    =
    \frac{e^{-t g}}{\int e^{-t g}\,dq}.
\]
Then
\[
    \KL(p_t\mid q)
    =
    \frac{t^2}{2}\operatorname{Var}_q[g]+o(t^2)
    \qquad\text{as }t\downarrow0.
\]
The same conclusion holds for \(p_t\) defined using tilts \(g_t\) in place of
\(g\), provided \(g_t\to g\) in \(L^2(q)\) and the family is locally uniformly
exponentially integrable.
\end{lem}

\begin{proof}
Let \(Z_t=\int e^{-t g}\,dq\). Since
\[
\KL(p_t\mid q)
=
\int \log\left(\frac{dp_t}{dq}\right)\,dp_t
=
-t\,\EE_{p_t}[g]-\log Z_t,
\]
the expansion follows from
\[
\log Z_t
=
-t\,\EE_q[g]
+
\frac{t^2}{2}\operatorname{Var}_q[g]
+
o(t^2),
\qquad
\EE_{p_t}[g]
=
\EE_q[g]-t\,\operatorname{Var}_q[g]+o(t).
\]
If \(g_t\to g\) in \(L^2(q)\) with local uniform exponential integrability, the
same Taylor expansions hold uniformly after replacing \(g\) by \(g_t\), and
\(\operatorname{Var}_q(g_t)\to\operatorname{Var}_q(g)\).
\end{proof}

\section{Stability}\label{sec:stability}

This section explains how the geometry of the gap loss converts sampling errors
and forward-model perturbations into parameter recovery guarantees. The proof
strategy is modular, so that the same argument can be reused for inverse UOT and
inverse JKO.

\subsection{A modular stability principle}

The first part of the section isolates the abstract mechanism behind all the
applications: once the observed moments, the empirical forward map, and the
curvature of the loss are controlled, the inverse parameter follows by a short
strong-convexity argument. In the previous section, we introduced
Fenchel--Young losses as a learning framework for variational models of the
form~\eqref{eq:forward}. Sharpened Fenchel--Young losses can be viewed as
ordinary Fenchel--Young losses associated with a modified functional. For this
reason, we state the stability theory below for an abstract empirical or
perturbed functional $\widehat\Omega_n$. In applications, $\widehat\Omega_n$ may
encode empirical plug-in estimates of auxiliary measures, sharpening terms, or
both.

We first record the abstract stability result. Its role is to isolate the three
properties that must be verified in applications: measurement stability, forward
stability, and local curvature.

\begin{thm}[Abstract stability of regularized gap minimization]
\label{thm:stability}
Let $\theta\in\RR^S$ and consider the parametrized family
\[
    V_\theta := \sum_{k=1}^S \theta_k \phi_k \in \Cc(\Zz).
\]
Let
\[
    \mu^\star = \mu_{\Omega}[V_{\theta^\star}]
\]
for some population functional $\Omega:\Mm_+(\Zz)\to \RR\cup\{+\infty\}$.
Fix $K>\|\theta^\star\|$. Let $\widehat\mu_n$ be an observed measure and let
$\widehat\Omega_n:\Mm_+(\Zz)\to \RR\cup\{+\infty\}$ be an empirical or perturbed
functional. Define
\[
    \widehat J_n(\theta)
    :=
    \mathcal L_{\widehat\Omega_n}^{\mathrm{FY}}(V_\theta;\widehat\mu_n)
    =
    \langle V_\theta,\widehat\mu_n\rangle
    -
    \inf_{\mu\in\Mm_+(\Zz)}
    \left\{
        \langle V_\theta,\mu\rangle+\widehat\Omega_n(\mu)
    \right\},
\]
and let
\[
    \widehat\theta_n
    \in
    \argmin_{\theta \in \mathbb{R}^S}\widehat J_n(\theta).
\]

Assume that there exist $\gamma_1,\gamma_2>0$ such that:
\begin{itemize}
    \item[(i)] \textbf{Measurement stability:}
    $
        \|
        \left(
        \langle \phi_i,\widehat\mu_n-\mu^\star\rangle
        \right)_{i=1}^S
        \|
        \leq \gamma_1.
    $

    \item[(ii)] \textbf{Forward stability.}
    $
        \|
        \left(
        \left\langle
        \phi_i,
        \mu_{\widehat\Omega_n}[V_{\theta^\star}]
        -
        \mu_{\Omega}[V_{\theta^\star}]
        \right\rangle
        \right)_{i=1}^S
        \|
        \leq \gamma_2.
    $

    \item[(iii)] \textbf{Local curvature.}
    $\widehat J_n$ is $\alpha$-strongly convex on $B_K(0)$.
\end{itemize}
If
\[
    \gamma_1+\gamma_2 < \alpha\,(K-\|\theta^\star\|),
\]
then the constrained estimator \(\widehat\theta_n\) is uniquely defined, and
\[
    \|\widehat\theta_n-\theta^\star\|
    \leq
    \frac{\gamma_1+\gamma_2}{\alpha}.
\]
\end{thm}

\begin{proof}
Write $V_\theta=\Phi\theta$, so that
\[
    \Phi^\ast \mu
    :=
    \left(
    \langle \phi_i,\mu\rangle
    \right)_{i=1}^S.
\]
Let
\[
    \widehat\theta_n
    \in
    \argmin_{\|\theta\|\leq K}\widehat J_n(\theta).
\]
We first show that the constraint is inactive. Since $\widehat J_n$ is
$\alpha$-strongly convex on $B_K(0)$,
\[
    \alpha\|\widehat\theta_n-\theta^\star\|^2
    \leq
    \left\langle
    \nabla\widehat J_n(\widehat\theta_n)
    -
    \nabla\widehat J_n(\theta^\star),
    \widehat\theta_n-\theta^\star
    \right\rangle.
\]
By optimality of $\widehat\theta_n$ over $B_K(0)$,
\[
    \left\langle
    \nabla\widehat J_n(\widehat\theta_n),
    \theta^\star-\widehat\theta_n
    \right\rangle
    \geq 0.
\]
Moreover,
\[
    \nabla\widehat J_n(\theta^\star)
    =
    \Phi^\ast \widehat\mu_n
    -
    \Phi^\ast \mu_{\widehat\Omega_n}[V_{\theta^\star}].
\]
Therefore,
\begin{align*}
    \alpha\|\widehat\theta_n-\theta^\star\|^2
    &\leq
    -
    \left\langle
    \Phi^\ast \widehat\mu_n
    -
    \Phi^\ast \mu_{\widehat\Omega_n}[V_{\theta^\star}],
    \widehat\theta_n-\theta^\star
    \right\rangle
    \\
    &=
    \left\langle
    \Phi^\ast\mu^\star-\Phi^\ast\widehat\mu_n,
    \widehat\theta_n-\theta^\star
    \right\rangle
    +
    \left\langle
    \Phi^\ast\mu_{\widehat\Omega_n}[V_{\theta^\star}]
    -
    \Phi^\ast\mu_{\Omega}[V_{\theta^\star}],
    \widehat\theta_n-\theta^\star
    \right\rangle
    \\
    &\leq
    (\gamma_1+\gamma_2)
    \|\widehat\theta_n-\theta^\star\|.
\end{align*}
Thus
\[
    \|\widehat\theta_n-\theta^\star\|
    \leq
    \frac{\gamma_1+\gamma_2}{\alpha}.
\]
The assumed strict inequality implies
\[
    \|\widehat\theta_n-\theta^\star\|<K-\|\theta^\star\|,
\]
hence $\|\widehat\theta_n\|<K$. Therefore the constrained minimizer lies in the
interior of $B_K(0)$ and satisfies the unconstrained first-order optimality
condition. Uniqueness follows from strong convexity on \(B_K(0)\).
\end{proof}

\paragraph{Comments on the assumptions.}

In applications, conditions (i)--(iii) are verified with high probability. The
measurement stability assumption is the most direct. Suppose that
$\mu^\star\in\Mm_+(\Zz)$ is observed through samples
\[
    z_1,\dots,z_n\overset{\mathrm{i.i.d.}}{\sim}\bar\mu^\star,
    \qquad
    \bar\mu^\star:=\frac{\mu^\star}{m_{\mu^\star}},
    \qquad
    m_{\mu^\star}:=\mu^\star(\Zz),
\]
and that the mass $m_{\mu^\star}$ is known. We use the convention
\[
    \widehat\mu_n
    :=
    m_{\mu^\star}
    \frac1n\sum_{i=1}^n\delta_{z_i}.
\]
Then
\[
    \left(
    \langle \phi_k,\widehat\mu_n-\mu^\star\rangle
    \right)_{k=1}^S
    =
    \frac{m_{\mu^\star}}{n}
    \sum_{i=1}^n
    \left(
    \phi_k(z_i)
    -
    \EE_{\bar\mu^\star}[\phi_k]
    \right)_{k=1}^S.
\]
If $\|(\phi_k(z))_{k=1}^S\|\leq C_\phi$ uniformly in $z$, then a vector
Bernstein inequality~\cite{gross2011recovering} gives, with probability at
least $1-t$,
\[
    \left\|
    \left(
    \langle \phi_k,\widehat\mu_n-\mu^\star\rangle
    \right)_{k=1}^S
    \right\|
    \leq
    C\,m_{\mu^\star} C_\phi
    \sqrt{\frac{\log(1/t)}{n}},
\]
up to universal constants.

The main work is usually to verify the forward stability and curvature
assumptions. The forward stability condition measures the sensitivity of the
forward map with respect to the perturbation
\[
    \Omega \longmapsto \widehat\Omega_n.
\]
The local curvature condition depends on the identifiable directions of the
parametrization. In many examples, the forward map has invariances. For example,
in optimal transport, replacing a cost $V(x,y)$ by
\[
    V(x,y)+f(x)+g(y)
\]
does not change the optimal coupling in the balanced case. Thus curvature can
only hold after removing such non-identifiable directions. This is reflected in
the choice of a suitable feature-centering operator, as in the inverse optimal
transport examples below.

\subsection{Application 1: Inverse unbalanced optimal transport}
\label{sec:stability-iuot}

We now apply Theorem~\ref{thm:stability} to inverse unbalanced optimal transport.
We use the empirical coupling and empirical reference marginals described in
Section~\ref{sec:examples_FY}:
\[
    \widehat\mu_n,
    \qquad
    \widehat\nu_1^n,
    \qquad
    \widehat\nu_2^n.
\]
For simplicity, the sample sizes of all three are assumed to be the same.

\paragraph{Affine cost parametrization.}

Consider the affine cost class
\[
    V_\theta(x,y)
    =
    \phi_0(x,y)+\sum_{k=1}^S\theta_k\phi_k(x,y).
\]
We assume that \(V^\star=V_{\theta^\star}\), and set
\[
    J(\theta)
    :=
    \mathcal L_{\Omega}^{\mathrm{FY}}(V_\theta;\mu^\star),
    \qquad
    \widehat J_n(\theta)
    :=
    \mathcal L_{\widehat\Omega_n}^{\mathrm{FY}}(V_\theta;\widehat\mu_n).
\]

\paragraph{Identifiability and feature design.}

A classical ambiguity in inverse optimal transport is that costs are identifiable
only up to additive terms of the form $f(x)+g(y)$. That is, in the balanced case,
$V(x,y)$ and $V(x,y)+f(x)+g(y)$ induce the same optimal coupling. In the
unbalanced setting, this ambiguity may be partially or fully removed by the
marginal divergence terms. The following feature decompositions isolate the
directions in which curvature can be expected.

Let $\phi=(\phi_1,\dots,\phi_S)$ denote the feature map. Define the population
marginal averages
\begin{align*}
    \phi^{(1)}(x)
    &:=
    \frac{1}{m_{\nu_2^\star}}
    \int_{\Yy}\phi(x,y)\,d\nu_2^\star(y),
    \\
    \phi^{(2)}(y)
    &:=
    \frac{1}{m_{\nu_1^\star}}
    \int_{\Xx}\phi(x,y)\,d\nu_1^\star(x),
    \\
    \phi^{(12)}
    &:=
    \frac{1}{m_{\nu_1^\star}m_{\nu_2^\star}}
    \int_{\Xx\times\Yy}\phi(x,y)\,
    d\nu_1^\star(x)\,d\nu_2^\star(y).
\end{align*}
Define the centered feature maps
\begin{equation}
\label{eq:feat_centred}
    \bar\phi^{(0)}
    :=
    \phi-\phi^{(1)}-\phi^{(2)}+\phi^{(12)},
    \qquad
    \bar\phi^{(1)}
    :=
    \phi-\phi^{(1)},
    \qquad
    \bar\phi^{(2)}
    :=
    \phi-\phi^{(2)}.
\end{equation}

We impose the following identifiability assumption. It states that one of the
centered feature covariances is nondegenerate in a direction where the transport
or marginal penalty supplies curvature.

\begin{assumption}
\label{ass:cost}
Assume that at least one of the following conditions holds:
\begin{itemize}
    \item[(i)]
    The matrix
    \[
        \left\langle
        \bar\phi^{(0)}(\bar\phi^{(0)})^\top,
        \bar\nu_1^\star\otimes\bar\nu_2^\star
        \right\rangle
        \in\RR^{S\times S}
    \]
    is invertible with smallest eigenvalue $\alpha_{\min}>0$.

    \item[(ii)]
    The matrix
    \[
        \left\langle
        \bar\phi^{(1)}(\bar\phi^{(1)})^\top,
        \bar\nu_1^\star\otimes\bar\nu_2^\star
        \right\rangle
        \in\RR^{S\times S}
    \]
    is invertible with smallest eigenvalue $\alpha_{\min}>0$, and
    $\varphi_2^\ast$ is locally strongly convex, locally Lipschitz, and smooth.

    \item[(iii)]
    The matrix
    \[
        \left\langle
        \phi\phi^\top,
        \bar\nu_1^\star\otimes\bar\nu_2^\star
        \right\rangle
        \in\RR^{S\times S}
    \]
    is invertible with smallest eigenvalue $\alpha_{\min}>0$, and both
    $\varphi_1^\ast$ and $\varphi_2^\ast$ are locally strongly convex, locally
    Lipschitz, and smooth.
\end{itemize}
Assume also that $\|\phi(x,y)\|\leq C_\phi$ uniformly on $\Xx\times\Yy$.
\end{assumption}

The three alternatives correspond to different degrees of marginal curvature.
Condition (i) is the balanced-type situation, where only the component of the
cost orthogonal to additive marginal terms is identifiable. Conditions (ii) and
(iii) use curvature in the marginal divergence terms to recover additional
directions.

\begin{thm}[Stability for inverse unbalanced optimal transport]
\label{thm:iuot-stability}
Work in the setting and notation above. Assume that \(\nu_1^\star\) and
\(\nu_2^\star\) are compactly supported, and that \(\Xx,\Yy\) are open subsets
of Euclidean space.
Let
\[
    \widehat\theta_n
    \in
    \argmin_{\theta \in \mathbb{R}^S}\widehat J_n(\theta).
\]
Assume that
\begin{itemize}
    \item \(\varphi_1^\ast\) and \(\varphi_2^\ast\) are locally Lipschitz
    smooth;
     \item 
There exist $a,a'\in\mathrm{dom}(\varphi_1)$ and $b,b'\in\mathrm{dom}(\varphi_2)$ such that
\[
bm_\beta > am_\alpha,
\qquad
b'm_\beta < a'm_\alpha.
\]
    \item either \(\varphi_1^\ast(s)=s\) and \(\varphi_2^\ast(s)=s\) are both
    identity functions, or at least one of \(\varphi_1^\ast,\varphi_2^\ast\) is
    locally strongly convex.
\end{itemize}
Let \(\alpha_{\min}>0\) be the smallest eigenvalue from
Assumption~\ref{ass:cost}, and set
\[
    K_\star\eqdef 2\|\theta^\star\|.
\]
If
\[
    n
    \ge
    C_0
    \left[
    \frac{C_\phi^4}{\alpha_{\min}^2}\bigl(t+\log(2S)\bigr)
    +
    \frac{C_\phi^2}{\alpha_{\min}}\bigl(t+\log(2S)+1\bigr)
    \right],
\]
for a sufficiently large constant \(C_0\), then, with probability at least
\(1-e^{-t}\), \(\widehat J_n\) is \(\alpha_\star\)-strongly convex on
\(B_{K_\star}(0)\), with
\[
    \alpha_\star
    \asymp
    \alpha_{\min}\exp(-C K_\star/\epsilon),
\]
and
\[
    \|\widehat\theta_n-\theta^\star\|
    \lesssim
    \frac{C_\phi}{\alpha_\star}
    \left(
    \sqrt{\frac{t+\log S}{n}}
    +
    \sqrt{S}\,
    e^{\|V^\star\|_\infty/\epsilon}
    \sqrt{\frac{\log n+t+\log S}{n}}
    \right),
\]
provided the right-hand side is smaller than \(\|\theta^\star\|\). If the
empirical marginals \(\widehat\nu_1^n\) and \(\widehat\nu_2^n\) are built from
independent samples, the \(\log n\) factor in the forward-stability term can be
removed.
\end{thm}

\begin{proof}
We apply Theorem~\ref{thm:stability} with
\[
    K=K_\star=2\|\theta^\star\|.
\]
First, the measurement term is controlled by bounded empirical concentration:
\[
    \left\|
    \left(
    \left\langle
    \phi_i,\widehat\mu_n-\mu^\star
    \right\rangle
    \right)_{i=1}^S
    \right\|
    \lesssim
    C_\phi\sqrt{\frac{t+\log S}{n}}.
\]
Second, Theorem~\ref{thm:sample-complexity-uot}, applied coordinatewise to
\(h=\phi_i\) and followed by a union bound, gives
\[
    \left\|
    \left(
    \left\langle
    \phi_i,
    \mu_{\widehat\Omega_n}[V_{\theta^\star}]
    -
    \mu_{\Omega}[V_{\theta^\star}]
    \right\rangle
    \right)_{i=1}^S
    \right\|
    \lesssim
    \sqrt{S}\,C_\phi
    e^{\|V^\star\|_\infty/\epsilon}
    \sqrt{\frac{\log n+t+\log S}{n}}.
\]
In the independent-marginal sampling case, the product-sampling refinement in
Theorem~\ref{thm:sample-complexity-uot} removes the factor \(\log n\).

It remains to verify the curvature hypothesis of Theorem~\ref{thm:stability}.
The stated lower bound on \(n\) implies
\[
    C C_\phi^2
    \left(
    \sqrt{\frac{t+\log(2S)}{n}}
    +
    \frac{t+\log(2S)}{n}
    +
    \frac1n
    \right)
    \leq
    c\,\alpha_{\min},
\]
after increasing \(C_0\) if needed. Therefore
Proposition~\ref{prop:strongConvexity} applies with \(B=K_\star\), and
\(\widehat J_n\) is \(\alpha_\star\)-strongly convex on \(B_{K_\star}(0)\).
The assumption that the displayed error bound is smaller than
\(K_\star-\|\theta^\star\|=\|\theta^\star\|\) is exactly the radius condition in
Theorem~\ref{thm:stability}. The theorem follows.
\end{proof}

\subsubsection{Forward stability}
Forward stability quantifies how empirical reference measures perturb the
coupling predicted by the true cost.
Given a cost \(V^\star\), consider the unbalanced OT coupling constructed from
population reference measures,
\[
\mu^\star
=
\argmin_{\mu\in\Mm_+(\Xx\times\Yy)}
\left\{
\int V^\star(x,y)\,d\mu(x,y)
+
\epsilon \KL(\mu\mid\nu_1^\star\otimes\nu_2^\star)
+
D_{\varphi_1}(\mu_1\mid\nu_1^\star)
+
D_{\varphi_2}(\mu_2\mid\nu_2^\star)
\right\},
\]
and the corresponding coupling constructed from empirical reference measures,
\[
\hat \mu^n
=
\argmin_{\mu\in\Mm_+(\Xx\times\Yy)}
\left\{
\int V^\star(x,y)\,d\mu(x,y)
+
\epsilon\KL(\mu\mid\hat \nu_1^n\otimes\hat \nu_2^n)
+
D_{\varphi_1}(\mu_1\mid\hat \nu_1^n)
+
D_{\varphi_2}(\mu_2\mid\hat \nu_2^n)
\right\}.
\]

We use the following result from~\cite{poon_UOT_complexity}.
\begin{thm}[Sample complexity for entropic unbalanced OT]\label{thm:sample-complexity-uot}
Assume that $\nu_1^\star, \nu_2^\star$ are compactly supported and that
$\Xx,\Yy$ are open subsets of Euclidean space. Let $\varphi_1,\varphi_2$ be convex functions defining \(\varphi\)-divergences. Assume that either  \(\varphi_1^*(x)=x\) and \(\varphi_2^*(x)=x\) are both
    identity functions (leading to balanced entropic optimal transport) or the following hold: 
\begin{itemize}
  \item  their convex conjugates \(\varphi_1^*, \varphi_2^*\) are locally Lipschitz smooth,
    \item  there exist $a,a'\in\mathrm{dom}(\varphi_1)$ and $b,b'\in\mathrm{dom}(\varphi_2)$ such that
\[
bm_\beta > am_\alpha,
\qquad
b'm_\beta < a'm_\alpha.
\] 
\item one of \(\varphi_1^*,\varphi_2^*\) is locally
    strongly convex.
\end{itemize}
Let \((z_i,w_i)_{i=1}^n \overset{\mathrm{i.i.d.}}{\sim} \xi\), where \(\xi\)
has marginals
\(\xi_1=\nu_1^\star/m_{\nu_1^\star}\) and
\(\xi_2=\nu_2^\star/m_{\nu_2^\star}\). Let
\[
\hat \nu_1^n := \frac{m_{\nu_1^\star}}{n}\sum_i \delta_{z_i},
\qquad
\hat \nu_2^n := \frac{m_{\nu_2^\star}}{n}\sum_i \delta_{w_i}.
\]

Then for all \(t>0\) and
\(h \in L^\infty(\nu_1^\star\otimes\nu_2^\star)\), with probability at least
\(1-e^{-t}\),
\begin{equation}\label{eq:samp-comp-forward}
\big| \langle h, \mu^\star-\hat \mu^n\rangle \big|
\;\lesssim\; e^{\|V^\star\|_\infty/\epsilon}
\|h\|_{L^\infty(\nu_1^\star\otimes \nu_2^\star)}
\sqrt{\frac{m_{\nu_1^\star} m_{\nu_2^\star}(\log n+t)}{n}}.
\end{equation}
Moreover, if
\(\xi=\frac{1}{m_{\nu_1^\star}m_{\nu_2^\star}}
\nu_1^\star\otimes\nu_2^\star\), the \(\log n\) factor can be removed.
\end{thm}

\subsubsection{Curvature}\label{sec:curvature_iuot}
Curvature is the mechanism that converts the forward-stability estimate into a
parameter-recovery statement for finite-dimensional costs. The argument below
shows how the unbalanced dual potentials generate a locally strongly convex
empirical objective after quotienting out the non-identifiable directions.

\begin{prop}[Local strong convexity]
\label{prop:strongConvexity}
Assume that $\Xx$ and $\Yy$ are compact and that Assumption~\ref{ass:cost}
holds. Fix $B>0$ and restrict to $\|\theta\|\leq B$. Assume that,
\[
    C C_\phi^2
    \left(
    \sqrt{\frac{\log(2S/\delta)}{n}}
    +
    \frac{\log(2S/\delta)}{n}
    +
    \frac1n
    \right)
    \leq
    c\,\alpha_{\min},
\]
for sufficiently small universal $c>0$. Then, with probability at least
$1-\delta$, the empirical objective $\widehat J_n$ is $\alpha$-strongly convex
on $\{\theta:\|\theta\|\leq B\}$, with
\[
    \alpha
    =
    C_1\,\alpha_{\min}\,\exp(-C_2B/\epsilon),
\]
where $C_1,C_2>0$ depend only on the feature bound, the masses, and the local
bounds on the dual potentials.
\end{prop}

\begin{proof}[Proof sketch]
The full argument is given in Appendix~\ref{app:proof-loc-conv}.
The proof is based on the dual formulation of the empirical UOT problem. By
convex duality,
\[
\begin{aligned}
    \widehat J_n(\theta)
    =
    \inf_{f,g}
    \Bigg\{
    &\langle V_\theta,\widehat\mu_n\rangle
    +
    \int_{\Xx}\varphi_1^\ast(-f)\,d\widehat\nu_1
    +
    \int_{\Yy}\varphi_2^\ast(-g)\,d\widehat\nu_2
    \\
    &+
    \epsilon
    \int_{\Xx\times\Yy}
    \exp\left(
    \frac{f(x)+g(y)-V_\theta(x,y)}{\epsilon}
    \right)
    d\widehat\nu_1(x)d\widehat\nu_2(y)
    \Bigg\}.
\end{aligned}
\]
Thus curvature of $\widehat J_n$ comes from the exponential term in the dual
objective, together with any local curvature of the marginal dual functions
$\varphi_1^\ast,\varphi_2^\ast$.

The only subtlety is that the exponential term is strongly convex in
\[
    f\oplus g - V_\theta,
\]
not directly in $V_\theta$. Therefore one must quotient out the additive
directions that can be absorbed into the dual potentials $f$ and $g$. In the
balanced case, this corresponds to the classical invariance
\[
    V(x,y)\sim V(x,y)+a(x)+b(y).
\]
Accordingly, we decompose the feature map into empirically centered components.

Under Assumption~\ref{ass:cost}(i), define
\[
    \widehat\psi^{(0)}(x,y)
    :=
    \phi(x,y)
    -
    \widehat\phi^{(1)}(x)
    -
    \widehat\phi^{(2)}(y)
    +
    \widehat\phi^{(12)}.
\]
Then $\widehat\psi^{(0)}$ is orthogonal to all additive functions
$a(x)+b(y)$ in
$L^2(\widehat\nu_1\otimes\widehat\nu_2)$. Hence the exponential term yields
curvature in the identifiable direction
\[
    (\theta-\theta')^\top\widehat\psi^{(0)}.
\]
More precisely, for
$\theta_t=t\theta+(1-t)\theta'$, one obtains
\[
    \widehat J_n(\theta_t)
    \leq
    t\widehat J_n(\theta)
    +(1-t)\widehat J_n(\theta')
    -
    c_\epsilon t(1-t)
    \left\|
    (\theta-\theta')^\top\widehat\psi^{(0)}
    \right\|^2_{L^2(\widehat\nu_1\otimes\widehat\nu_2)},
\]
where
\[
    c_\epsilon\gtrsim \epsilon^{-1}\exp(-C B/\epsilon).
\]
Finally, Assumption~\ref{ass:cost}(i), together with the empirical covariance
concentration estimate, implies
\[
    \left\|
    (\theta-\theta')^\top\widehat\psi^{(0)}
    \right\|^2_{L^2(\widehat\nu_1\otimes\widehat\nu_2)}
    \geq
    c\,\alpha_{\min}\|\theta-\theta'\|^2
\]
with high probability.

Under Assumption~\ref{ass:cost}(ii), one marginal dual term, say
$\varphi_2^\ast$, has local strong convexity. This curvature controls one
additive direction, so only the remaining $x$-dependent component must be
removed. The relevant feature map is the one-sided centering
\[
    \widehat\psi^{(1)}(x,y)
    :=
    \phi(x,y)-\widehat\phi^{(1)}(x).
\]
The curvature of $\varphi_2^\ast$ is combined with the exponential curvature to
control the cross term between the dual variable $g$ and
$(\theta-\theta')^\top\widehat\psi^{(1)}$. This gives curvature proportional to
the empirical covariance of $\widehat\psi^{(1)}$, which is again bounded below
by Assumption~\ref{ass:cost}(ii) and a concentration estimate.

Finally, in the case of Assumption~\ref{ass:cost}(iii), both marginal dual
terms are locally strongly convex. The marginal curvature controls both
additive directions, so no centering is needed; curvature is controlled by the
empirical covariance of the original feature map $\phi$.


\end{proof}

\subsection{Application 2: Inverse gradient flow}
\label{sec:stability-ijko}

We now consider the setting of Section~\ref{sec:sharpened_iGF}. The goal is to
recover a potential function $V^\star \in \Cc(\Zz)$ given i.i.d.\ samples from
two consecutive snapshots \(\nu_1^\star,\nu_2^\star\in\Pp(\Zz)\) satisfying
\[
\nu_2^\star
=
\argmin_{\mu\in\Pp(\Zz)}
\left\{
\int_{\Zz} V^\star(z)\,d\mu(z)
+
\frac{1}{\tau} W_{2,\epsilon}^2(\mu,\nu_1^\star)
\right\}.
\]
Here \(\nu_1^\star\) is the previous snapshot and \(\nu_2^\star\) is the next
snapshot. The observed measure in the abstract stability theorem is therefore
\(\mu^\star=\nu_2^\star\). Given empirical snapshots
\(\hat\nu_1^n,\hat\nu_2^n\), we use the sharpened empirical functional
\[
\widehat\Omega_n(\mu)
\eqdef
\frac{1}{\tau}W_{2,\epsilon}^2(\mu,\hat\nu_1^n)
+
r\,\KL(\mu\mid\hat\nu_2^n),
\qquad \mu\in\Pp(\Zz).
\]
Throughout this subsection we assume
\[
    r' \eqdef r-\epsilon/\tau \ge 0.
\]
The role of this condition is only to ensure that, after the reduction below,
the induced marginal penalty is again a nonnegative KL divergence.

\begin{thm}[Stability for inverse gradient flow]
\label{thm:ijko-stability}
Assume that \(\Zz\subset\RR^d\) is compact and that
\[
    V^\star=V_{\theta^\star},
    \qquad
    V_\theta(z)=\sum_{i=1}^S\theta_i\phi_i(z),
    \qquad
    \|\phi(z)\|\le C_\phi.
\]
Let \(z_1^1,\ldots,z_n^1\) be i.i.d.\ samples from \(\nu_1^\star\), let
\(z_1^2,\ldots,z_n^2\) be i.i.d.\ samples from \(\nu_2^\star\), and assume the
two samples are independent of one another. Let \(\hat\nu_1^n\) and
\(\hat\nu_2^n\) be the corresponding empirical measures. 
Let
\[
    \widehat\theta_n
    \in
    \argmin_{\theta \in \mathbb{R}^S}
    \widehat J_n(\theta),
    \qquad
    \widehat J_n(\theta)
    \eqdef
    \left\langle V_\theta,\hat\nu_2^n\right\rangle
    -
    \inf_{\mu\in\Pp(\Zz)}
    \left\{
    \left\langle V_\theta,\mu\right\rangle
    +
    \widehat\Omega_n(\mu)
    \right\}.
\]
Let
\[
    C_{V^\star}(y,x)
    \eqdef
    V^\star(x)+\tau^{-1}\|x-y\|^2.
\]
Assume that \(r'>0\) and that the centered covariance
\[
    \mathbf M
    \eqdef
    \EE_{z\sim\nu_2^\star}
    \left[
    \bar\phi(z)\bar\phi(z)^\top
    \right],
    \qquad
    \bar\phi(z)=\phi(z)-\EE_{\nu_2^\star}[\phi],
\]
has smallest eigenvalue \(\alpha_{\min}>0\).
If
\[
    n
    \ge
    C_0
    \left[
    \frac{C_\phi^4}{\alpha_{\min}^2}\bigl(t+\log(2S)\bigr)
    +
    \frac{C_\phi^2}{\alpha_{\min}}\bigl(t+\log(2S)+1\bigr)
    \right],
\]
for a sufficiently large constant \(C_0\), then, with probability at least
\(1-e^{-t}\), \(\widehat J_n\) is \(\alpha_\star\)-strongly convex on
\(B_{K_\star}(0)\), with
\[
    \alpha_\star
    \asymp
    \alpha_{\min}
    \exp\!\left(-\frac{C K_\star}{\epsilon/\tau}\right)
    \quad\text{and}\quad
    K_\star \eqdef 2\|\theta^\star\|.
\]
and
\[
    \|\widehat\theta_n-\theta^\star\|
    \lesssim
    \frac{C_\phi}{\alpha_\star}
    \left(
    1
    +
    \sqrt{S}\,
    \exp\!\left(\frac{\|C_{V^\star}\|_\infty}{\epsilon/\tau}\right)
    \right)
    \sqrt{\frac{t+\log S}{n}},
\]
provided the right-hand side is smaller than \(\|\theta^\star\|\).
\end{thm}

\begin{proof}
We apply Theorem~\ref{thm:stability} with
\[
    \mu^\star=\nu_2^\star,
    \qquad
    \widehat\mu_n=\hat\nu_2^n,
    \qquad
    K=K_\star=2\|\theta^\star\|.
\]
The measurement term satisfies, by standard bounded empirical concentration and
a union bound over the \(S\) coordinates,
\[
    \left\|
    \left(
    \left\langle
    \phi_i,\hat\nu_2^n-\nu_2^\star
    \right\rangle
    \right)_{i=1}^S
    \right\|
    \lesssim
    C_\phi
    \sqrt{\frac{t+\log S}{n}}.
\]
The forward-stability term is controlled by Proposition below, applied to each
coordinate \(h=\phi_i\), followed by a union bound. Indeed, the reduction to
iUOT has a hard constraint on the first marginal and a KL divergence on the
second marginal. The corresponding conjugates are locally Lipschitz and smooth,
and the KL conjugate is locally strongly convex because \(r'>0\). Hence the
hypotheses of Theorem~\ref{thm:sample-complexity-uot} hold for the induced
iUOT problem. Since the two snapshot samples are independent, we are in the
product-sampling case and the logarithmic factor in \(n\) is absent:
\[
    \left\|
    \left(
    \left\langle
    \phi_i,
    \mu_{\widehat\Omega_n}[V_{\theta^\star}]
    -
    \mu_{\Omega}[V_{\theta^\star}]
    \right\rangle
    \right)_{i=1}^S
    \right\|
    \lesssim
    \sqrt{S}\,C_\phi
    \exp\!\left(\frac{\|C_{V^\star}\|_\infty}{\epsilon/\tau}\right)
    \sqrt{\frac{t+\log S}{n}}.
\]
It remains to verify the curvature hypothesis. The lower bound on \(n\) implies
\[
    C C_\phi^2
    \left(
    \sqrt{\frac{t+\log(2S)}{n}}
    +
    \frac{t+\log(2S)}{n}
    +
    \frac1n
    \right)
    \le
    c\,\alpha_{\min},
\]
after increasing \(C_0\) if necessary. Thus Proposition~\ref{prop:strongConvexity}
applies to the induced iUOT problem on \(B_{K_\star}(0)\), and the curvature
reduction below transfers this strong convexity to \(\widehat J_n\) with
constant \(\alpha_\star\). The assumption that the displayed error bound is
smaller than \(K_\star-\|\theta^\star\|=\|\theta^\star\|\) is exactly the
radius condition in Theorem~\ref{thm:stability}. Applying
Theorem~\ref{thm:stability} gives the result.
\end{proof}

\paragraph{Reduction to iUOT.}
For any \(V\in\Cc(\Zz)\), define the cost on \(\Zz\times\Zz\)
\[
    C_V(y,x)
    \eqdef
    V(x)+\frac1\tau\|x-y\|^2,
\]
where the first coordinate \(y\) corresponds to the previous snapshot and the
second coordinate \(x\) corresponds to the next snapshot. Consider the inner
problem
\[
A_n(V)
\eqdef
\inf_{\mu\in\Pp(\Zz)}
\left\{
\langle V,\mu\rangle
+
\frac{1}{\tau}W_{2,\epsilon}^2(\mu,\hat\nu_1^n)
+
r\,\KL(\mu\mid\hat\nu_2^n)
\right\}.
\]
If \(\KL(\mu\mid\hat\nu_2^n)<\infty\), then \(\mu\ll\hat\nu_2^n\). Hence, for
any coupling \(\pi\) with first marginal \(\hat\nu_1^n\) and second marginal
\(\mu\), the chain rule for relative entropy gives
\[
\KL(\pi\mid\hat\nu_1^n\otimes\mu)
=
\KL(\pi\mid\hat\nu_1^n\otimes\hat\nu_2^n)
-
\KL(\mu\mid\hat\nu_2^n).
\]
Expanding the entropic transport term and using the identity above yields
\begin{equation}\label{eq:JKO_TO_UOT}
\begin{aligned}
A_n(V)
=
\inf_{\pi\in\Pp(\Zz\times\Zz)}
\Bigg\{
&\int_{\Zz\times\Zz} C_V(y,x)\,d\pi(y,x)
+
\frac{\epsilon}{\tau}
\KL(\pi\mid\hat\nu_1^n\otimes\hat\nu_2^n)
\\
&\qquad
+
r'\,\KL(\pi_2\mid\hat\nu_2^n)
+
\iota_{\{\pi_1=\hat\nu_1^n\}}(\pi)
\Bigg\}.
\end{aligned}
\end{equation}
Thus the sharpened iJKO inner problem is an unbalanced OT problem with
entropic parameter \(\epsilon/\tau\), a hard constraint on the previous-snapshot
marginal, and a KL penalty on the next-snapshot marginal. The same identity
holds at the population level after replacing
\((\hat\nu_1^n,\hat\nu_2^n)\) by \((\nu_1^\star,\nu_2^\star)\).

\subsubsection{Forward stability}
The JKO stability estimate follows by reading the sharpened inner problem as an
unbalanced OT problem and then projecting the transport stability bound onto the
next-snapshot marginal. This is the point where the KL-sharpened JKO loss
inherits the finite-sample behavior of the induced UOT problem.

Let
\[
\mu_{\Omega}(V)
\in
\argmin_{\mu\in\Pp(\Zz)}
\left\{
\langle V,\mu\rangle+\frac1\tau W_{2,\epsilon}^2(\mu,\nu_1^\star)
\right\},
\]
and let \(\mu_{\widehat\Omega_n}(V)\) be the corresponding minimizer with
\(\widehat\Omega_n\). Let \(\pi_{V^\star}\) and \(\pi_{V^\star}^n\) denote the
population and empirical iUOT minimizers in the
representation~\eqref{eq:JKO_TO_UOT} with \(V=V^\star\). By calibration of the
sharpened loss,
\[
    \mu_{\Omega}(V^\star)=\nu_2^\star=(\pi_{V^\star})_2,
    \qquad
    \mu_{\widehat\Omega_n}(V^\star)=(\pi_{V^\star}^n)_2.
\]

\begin{prop}[Forward stability for iJKO via iUOT]
\label{prop:ijko-forward-stability}
Assume that the hypotheses of Theorem~\ref{thm:sample-complexity-uot} hold for
the induced iUOT problem~\eqref{eq:JKO_TO_UOT} with cost \(C_{V^\star}\),
reference marginals \((\nu_1^\star,\nu_2^\star)\), entropic parameter
\(\epsilon/\tau\), and divergences
\[
    D_{\varphi_1}(\cdot\mid\nu_1^\star)=\iota_{\{\cdot=\nu_1^\star\}},
    \qquad
    D_{\varphi_2}(\cdot\mid\nu_2^\star)=r'\KL(\cdot\mid\nu_2^\star).
\]
Then, for every \(h\in L^\infty(\nu_2^\star)\), with probability at least
\(1-e^{-t}\),
\[
    \left|
    \left\langle
    h,\mu_{\widehat\Omega_n}(V^\star)-\mu_{\Omega}(V^\star)
    \right\rangle
    \right|
    \lesssim
    \|h\|_\infty
    \exp\!\left(\frac{\|C_{V^\star}\|_\infty}{\epsilon/\tau}\right)
    \sqrt{\frac{\log n+t}{n}}.
\]
If the samples from \(\nu_1^\star\) and \(\nu_2^\star\) are independent, the
\(\log n\) factor can be removed under the product-sampling case of
Theorem~\ref{thm:sample-complexity-uot}.
\end{prop}
\begin{proof}
The reduction above identifies the forward predictions
\(\mu_{\Omega}(V^\star)\) and \(\mu_{\widehat\Omega_n}(V^\star)\) with the
second marginals of the population and empirical iUOT optimizers,
respectively. For \(h\in L^\infty(\nu_2^\star)\), define
\[
    \widetilde h(y,x)\eqdef h(x).
\]
Then
\[
\left\langle
h,\mu_{\widehat\Omega_n}(V^\star)-\mu_{\Omega}(V^\star)
\right\rangle
=
\left\langle
\widetilde h,\pi_{V^\star}^n-\pi_{V^\star}
\right\rangle.
\]
Applying Theorem~\ref{thm:sample-complexity-uot} to the induced iUOT problem
with test function \(\widetilde h\) gives the claimed bound.
\end{proof}

\subsubsection{Curvature}
The final ingredient is a curvature estimate for the empirical sharpened iJKO
objective, obtained by transferring the iUOT curvature result through the same
reduction.
\begin{prop}[Curvature for empirical iJKO via iUOT]
\label{prop:ijko-curvature}
Let
\[
    \widehat\Jj_n(\theta)
    \eqdef
    \left\langle V_\theta,\hat\nu_2^n\right\rangle
    -
    \inf_{\mu\in\Pp(\Zz)}
    \left\{
    \left\langle V_\theta,\mu\right\rangle
    +
    \widehat\Omega_n(\mu)
    \right\},
    \qquad
    V_\theta(z)=\sum_{i=1}^S\theta_i\phi_i(z).
\]
Assume that \(\Zz\) is compact, that \(r'>0\), and that
\(\|\phi(z)\|\le C_\phi\). Assume also that the centered feature covariance
\[
\mathbf{M}
\;\eqdef\;
\EE_{z\sim\nu_2^\star}\!\big[\bar\phi(z)\bar\phi(z)^\top\big],
\qquad
\bar\phi(z)=\phi(z)-\EE_{\nu_2^\star}[\phi],
\]
is invertible with smallest eigenvalue \(\alpha_{\min}>0\). Fix \(B>0\). If
\[
    C C_\phi^2
    \left(
    \sqrt{\frac{t+\log(2S)}{n}}
    +
    \frac{t+\log(2S)}{n}
    +
    \frac1n
    \right)
    \le
    c\,\alpha_{\min},
\]
for the constants in Proposition~\ref{prop:strongConvexity}, then with
probability at least \(1-e^{-t}\), \(\widehat\Jj_n\) is strongly convex on
\(\{\theta:\|\theta\|\le B\}\), with constant
\[
    \alpha
    \asymp
    \alpha_{\min}
    \exp\!\left(-\frac{C B}{\epsilon/\tau}\right),
\]
where \(C\) depends on the feature bound, the support, \(r'\), and the local
dual bounds in the induced iUOT problem.
\end{prop}
\begin{proof}
By~\eqref{eq:JKO_TO_UOT}, up to constants independent of \(\theta\),
\(\widehat\Jj_n\) is the empirical iUOT Fenchel--Young loss associated with
the affine cost family
\[
    C_\theta(y,x)
    =
    \frac1\tau\|x-y\|^2+\sum_{i=1}^S\theta_i\phi_i(x).
\]
The fixed transport term \(\tau^{-1}\|x-y\|^2\) is part of the baseline cost and
does not affect the parameter Hessian. The parameter features for the induced
iUOT problem are
\[
    \psi_i(y,x)\eqdef \phi_i(x).
\]
Since the first marginal is constrained to equal \(\hat\nu_1^n\) and the KL
penalty \(r'\KL(\pi_2\mid\hat\nu_2^n)\) acts on the second marginal, the
problem falls under Assumption~\ref{ass:cost}(ii) for iUOT.\@ Indeed, for
\(\psi=(\psi_1,\ldots,\psi_S)\),
\[
    \psi^{(1)}(y)
    =
    \int_{\Zz}\psi(y,x)\,d\nu_2^\star(x)
    =
    \EE_{\nu_2^\star}[\phi],
    \qquad
    \bar\psi^{(1)}(y,x)
    =
    \phi(x)-\EE_{\nu_2^\star}[\phi].
\]
Thus the covariance required in Assumption~\ref{ass:cost}(ii) is exactly
\(\mathbf M\), up to the mass of \(\nu_1^\star\), which is one here. The
empirical concentration condition above ensures that the corresponding
empirical covariance remains bounded below by a constant multiple of
\(\alpha_{\min}\). Applying Proposition~\ref{prop:strongConvexity} to the
induced iUOT problem, with effective entropic parameter \(\epsilon/\tau\), gives
the claimed strong convexity bound for \(\widehat\Jj_n\).
\end{proof}

\section{Numerical experiments}\label{sec:numerics}
These experiments check whether the geometric advantages of sharpening are
visible in finite computations, both through improved conditioning and through
more reliable sparse recovery from sampled trajectories.

\subsection{Density estimation}
We illustrate the conditioning effect of sharpening in the precision-estimation
example of Section~\ref{sec:precision_sharpening_example}. We implement the
sharpened Fenchel--Young loss after replacing the continuous reference measure
\(dx\) by a finite empirical reference measure.

\paragraph{Empirical setup.}
Let \(z_1,\ldots,z_N\in\RR^d\) be reference particles sampled uniformly from a
box, and set
\[
    \rho_N \eqdef \frac1N\sum_{i=1}^N\delta_{z_i}.
\]
This is the empirical substitute for \(dx\) in the entropic density-estimation
functional. We consider diagonal precision matrices
\[
    K_\lambda=\mathrm{diag}(\lambda_1,\ldots,\lambda_d),
    \qquad
    V_\lambda(z)=\frac12\sum_{j=1}^d\lambda_j z_j^2,
\]
and take the calibrated precision to be \(\lambda^\star=\mathbf 1\). The
observed empirical law is the discrete Gibbs measure on the same support,
\[
    \mu_N^\star
    =
    \sum_{i=1}^N p_i^\star\delta_{z_i},
    \qquad
    p_i^\star
    =
    \frac{\exp(-V_{\lambda^\star}(z_i))}
    {\sum_{k=1}^N\exp(-V_{\lambda^\star}(z_k))}.
\]
For \(r\ge0\), the empirical sharpened inner problem is the finite-dimensional
simplex problem
\[
    \inf_{q\in\Delta_N}
    \left\{
    \sum_{i=1}^N q_i V_\lambda(z_i)
    +
    \KL(q\mid \rho_N)
    +
    r\,\KL(q\mid \mu_N^\star)
    \right\}.
\]
Up to constants independent of \(q\), its optimizer is explicit:
\[
    q_i^{\lambda,r}
    =
    \frac{
    \exp\left(
    \frac{-V_\lambda(z_i)+r\log p_i^\star}{1+r}
    \right)
    }
    {
    \sum_{k=1}^N
    \exp\left(
    \frac{-V_\lambda(z_k)+r\log p_k^\star}{1+r}
    \right)
    }.
\]
Dropping constants independent of \(\lambda\), the empirical sharpened
Fenchel--Young objective is
\[
    \widehat J_r(\lambda)
    =
    \sum_{i=1}^N p_i^\star V_\lambda(z_i)
    +
    (1+r)
    \log
    \sum_{i=1}^N
    \exp\left(
    \frac{-V_\lambda(z_i)+r\log p_i^\star}{1+r}
    \right).
\]
Writing
\[
    \phi(z)=\frac12(z_1^2,\ldots,z_d^2),
\]
we have
\[
    \nabla \widehat J_r(\lambda)
    =
    \EE_{\mu_N^\star}[\phi]
    -
    \EE_{q^{\lambda,r}}[\phi],
    \qquad
    \nabla^2 \widehat J_r(\lambda)
    =
    \frac1{1+r}\operatorname{Cov}_{q^{\lambda,r}}(\phi).
\]
Thus the numerical experiment evaluates the same empirical gap geometry as the
sharpened loss, with all integrals replaced by finite sums over the reference
particles.

\begin{figure}[htp]
    \centering
    \includegraphics[width=0.96\textwidth]{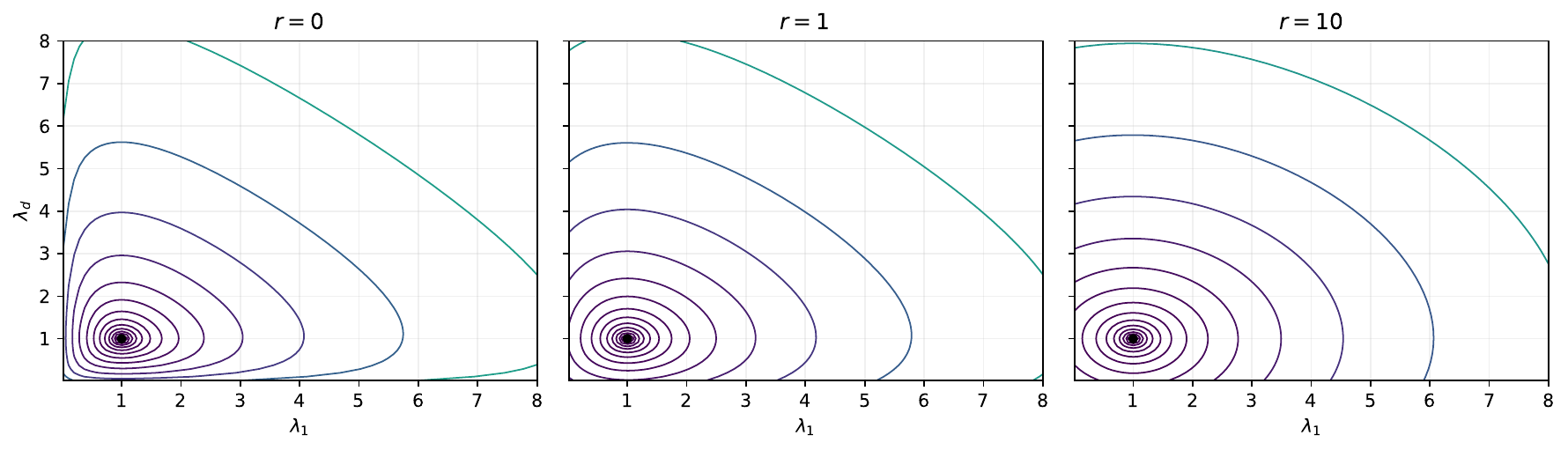}
    \caption{
    Two-dimensional slices of the empirical sharpened Fenchel--Young loss
    \(\widehat J_r(\lambda)-\widehat J_r(\lambda^\star)\). We vary
    \((\lambda_1,\lambda_d)\) and keep all other coordinates fixed at
    \(\lambda_j^\star=1\). The black dot marks the calibrated precision.
    }
    \label{fig:precision-loss-contours}
\end{figure}

\begin{figure}[htp]
    \centering
    \subfigure[Condition number at initialization.]{
        \includegraphics[width=0.47\textwidth]{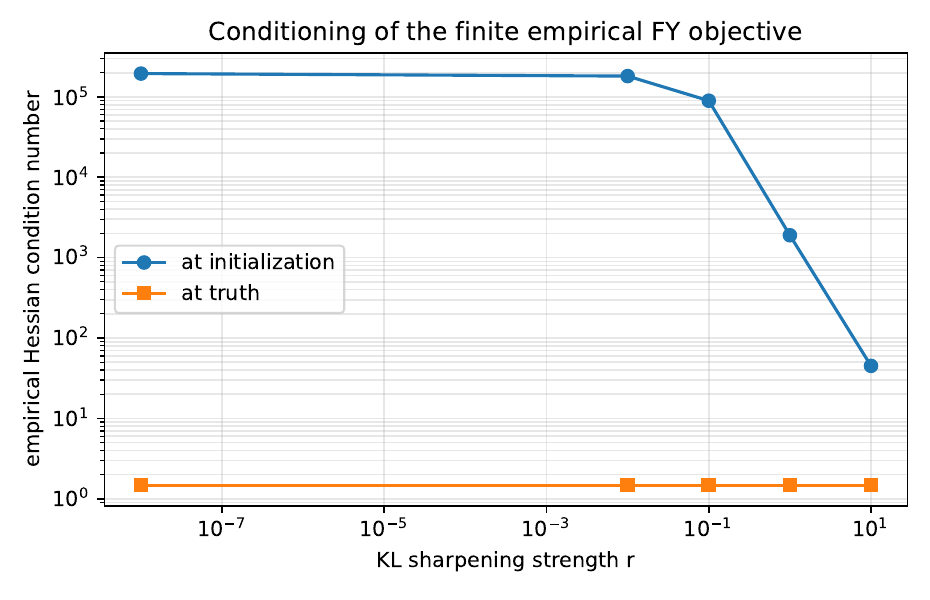}
    }
    \hfill
    \subfigure[Gradient descent from the ill-conditioned initialization.]{
        \includegraphics[width=0.47\textwidth]{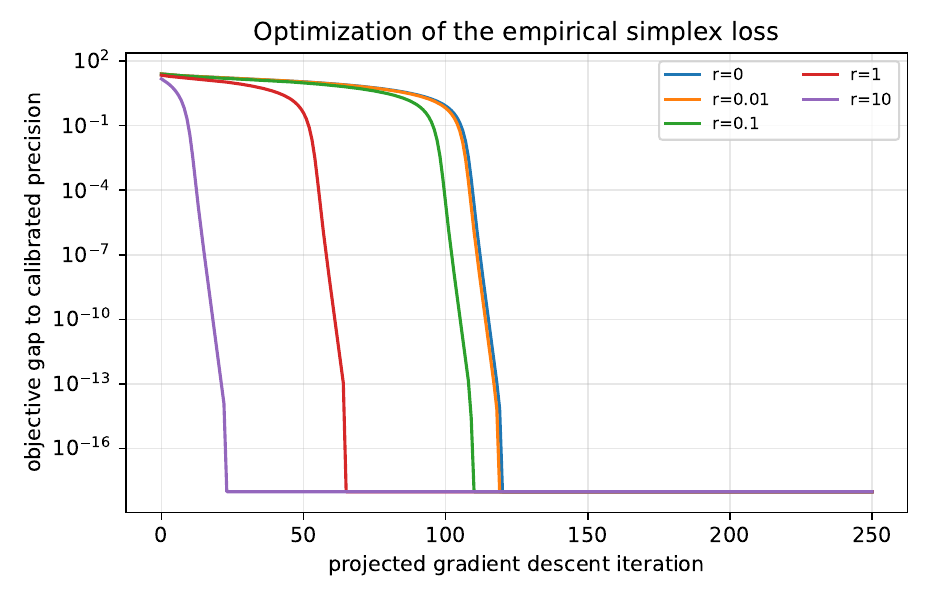}
    }
    \caption{
    Sharpening improves the empirical optimization geometry away from the
    calibrated precision. Left: the condition number of
    \(\nabla^2\widehat J_r(\lambda)\) at the ill-conditioned initialization
    decreases as \(r\) grows. Right: the corresponding gradient-descent traces
    show faster decrease for the better-conditioned sharpened objectives.
    }
    \label{fig:precision-conditioning-gd}
\end{figure}

We use \(N=20000\), \(d=5\), reference particles sampled uniformly on
\([-6,6]^d\), and compare \(r\in\{0,10^{-2},10^{-1},1,10\}\). The gradient
descent initialization is an ill-conditioned diagonal precision with spectrum
geometrically spaced in \([0.02,50]\). Figure~\ref{fig:precision-loss-contours}
shows two-dimensional loss contours as \(r\) varies: increasing \(r\) improves
the curvature away from the minimum. This geometric effect is reflected in
optimization. Figure~\ref{fig:precision-conditioning-gd} shows that increasing
\(r\) improves the conditioning at initialization and accelerates gradient
descent.

\subsection{Inverse gradient flow}

We next illustrate the inverse-gradient-flow example of
Section~\ref{sec:sharpened_iGF}. The experiment uses the semidual form of the
empirical sharpened iJKO loss, which avoids optimizing directly over the next
snapshot measure at every value of the parameter.

\paragraph{Semidual formulation.}
Consider two consecutive empirical snapshots, written with the next snapshot
first,
\[
    \hat\nu_{k+1}=\frac1M\sum_{i=1}^M\delta_{y_i},
    \qquad
    \hat\nu_k=\frac1N\sum_{j=1}^N\delta_{x_j},
\]
and define
\[
    \eta_\tau\eqdef \epsilon/\tau,
    \qquad
    s\eqdef r-\eta_\tau>0,
    \qquad
    c_\theta(y_i,x_j)
    \eqdef
    V_\theta(y_i)+\frac1\tau\|y_i-x_j\|^2.
\]
The semidual used in the computations follows from the Kantorovich
formulation of the sharpened iJKO loss. After eliminating the dual potential on
the previous snapshot, the empirical loss for this pair of snapshots is
\[
\widehat{\mathcal L}_{\mathrm{iJKO},r}^{\sharp}
(V_\theta;\hat\nu_k,\hat\nu_{k+1})
    =
\min_{f\in\RR^M} S_k(\theta,f),
\]
where
\[
\begin{aligned}
S_k(\theta,f)
&\eqdef
\frac1M\sum_{i=1}^M V_\theta(y_i)
+
\frac{s}{M}\sum_{i=1}^M \exp\left(-\frac{f_i}{s}\right)
\\
&\quad
+
\frac1N\sum_{j=1}^N
\eta_\tau
\log\left[
\frac1M
\sum_{i=1}^M
\exp\left(
\frac{f_i-c_\theta(y_i,x_j)}{\eta_\tau}
\right)
\right].
\end{aligned}
\]
Equivalently, written before discretization,
\[
\begin{aligned}
S_k(V,f)
&=
\left\langle
V+s\exp\left(-\frac{f}{s}\right),
\nu_{k+1}
\right\rangle
\\
&\quad
+
\int
\eta_\tau
\log
\left(
\int
\exp
\left(
\frac{f(y)-V(y)-\tau^{-1}\|y-x\|^2}{\eta_\tau}
\right)
d\nu_{k+1}(y)
\right)
d\nu_k(x).
\end{aligned}
\]
Thus the aggregate empirical objective is
\[
\widehat J_r(\theta)
    =
    \sum_{k=0}^{T-1}
\min_{f^k\in\RR^{M_k}} S_k(\theta,f^k).
\]

\paragraph{Gaussian quadratic setup.}
We consider quadratic potentials
\[
    V_\theta(x)=x^\top\theta x,
    \qquad
    \theta=\theta^\top\in\RR^{d\times d}.
\]
The data are generated by the linear gradient flow
\[
    \frac{d}{dt}x_t=-\nabla V_{\theta^\star}(x_t),
    \qquad
    x_0\sim\Nn(m^\star,\Sigma^\star).
\]
Hence each snapshot remains Gaussian:
\[
    x_t\sim\Nn(m_t,\Sigma_t),
    \qquad
    m_t=e^{-2t\theta^\star}m^\star,
    \qquad
    \Sigma_t=e^{-2t\theta^\star}\Sigma^\star e^{-2t\theta^\star}.
\]
For \(k=0,\ldots,T\), we draw \(N\) independent samples from the law at time
\(\tau k\). The estimator is obtained by minimizing the regularized objective
\[
    \widehat\theta_{\lambda}
    \in
    \argmin_{\theta=\theta^\top,\ \theta\ge0}
    \left\{
    \widehat J_r(\theta)+\lambda\|\theta\|_1
    \right\},
\]
where the constraint \(\theta\ge0\) is imposed entrywise in the sparse graph
experiment. We fix \(r=1\) in the numerical tests. As
in~\cite{andradesparsistency}, the semidual formulation with \(\ell_1\)
regularization can be optimized by BFGS after replacing \(\norm{\theta}_1\) by
its quadratic variational form.

\begin{figure}[htp]
    \centering
    \begin{tabular}{ccc}
        \includegraphics[height=0.22\linewidth]{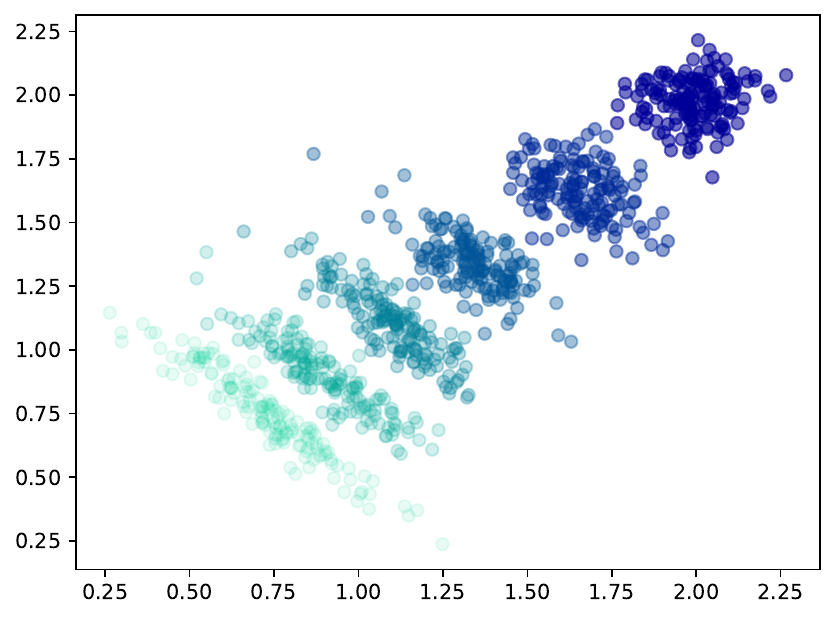}
        &
        \includegraphics[height=0.22\linewidth]{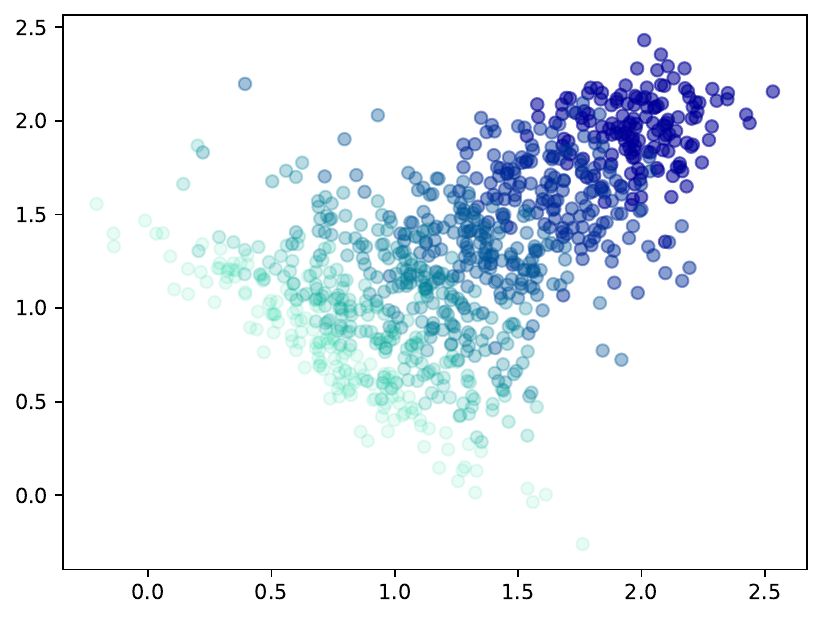}
        &
        \includegraphics[height=0.22\linewidth]{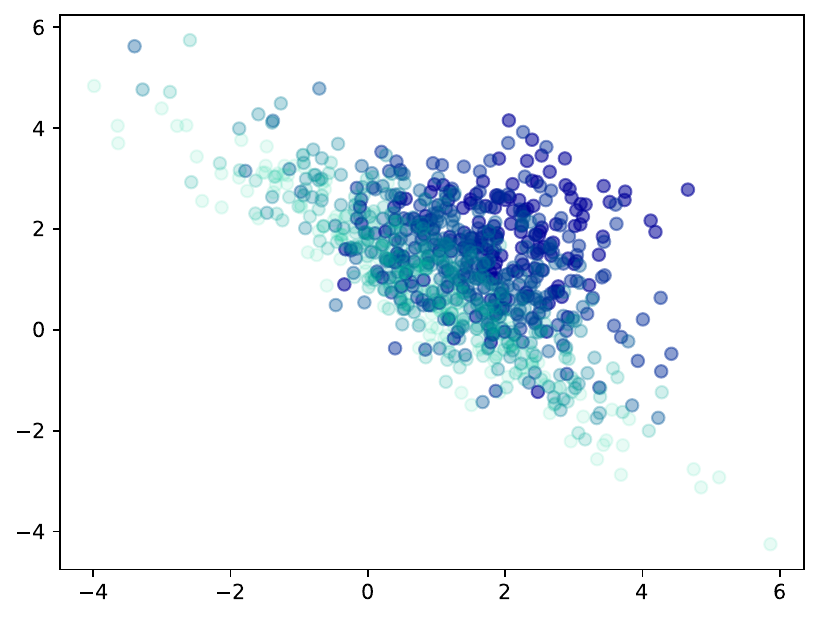}
        \\
        \(\sigma=0.1\)& \(\sigma=0.2\)&\(\sigma=1\)
    \end{tabular}
    \caption{
    Evolution across six time points for the sparse inverse-gradient-flow
    experiment, with \(\tau=0.1\), \(m^\star=2\mathbf 1\), and
    \(\Sigma^\star=\sigma^2\Id\).
    }
    \label{fig:trajectory-l1}
\end{figure}

\begin{figure}[htp]
    \centering
    \begin{tabular}{ccc}
        \(\sigma=0.1, T=2\)
        &
        \(\sigma=0.1, T=6\)
        &
        \(\sigma=1, T=2\)
        \\
        \includegraphics[width=0.3\linewidth]{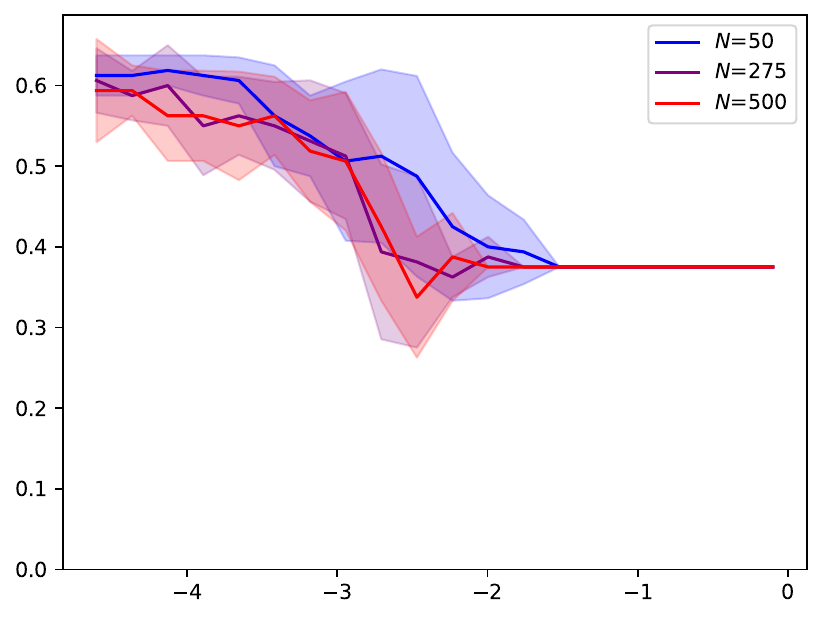}
        &
        \includegraphics[width=0.3\linewidth]{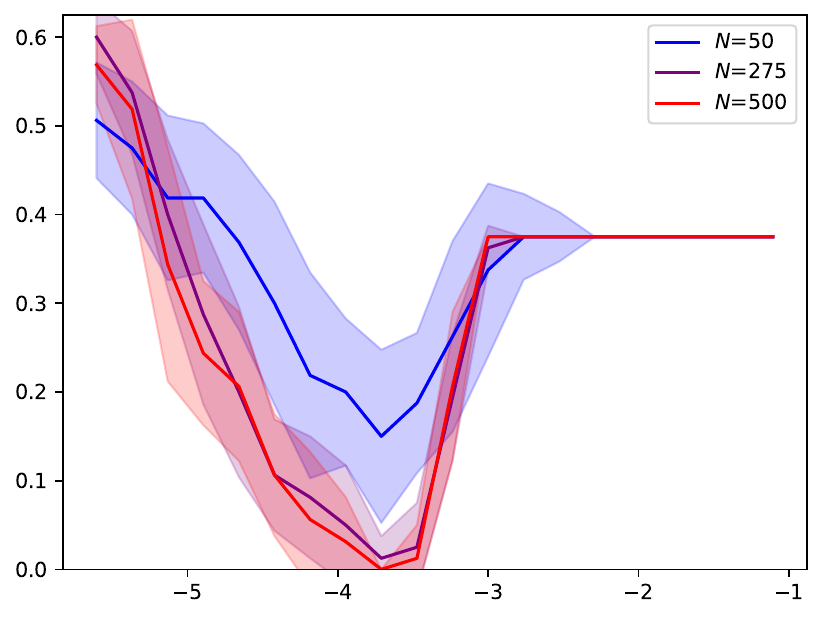}
        &
        \includegraphics[width=0.3\linewidth]{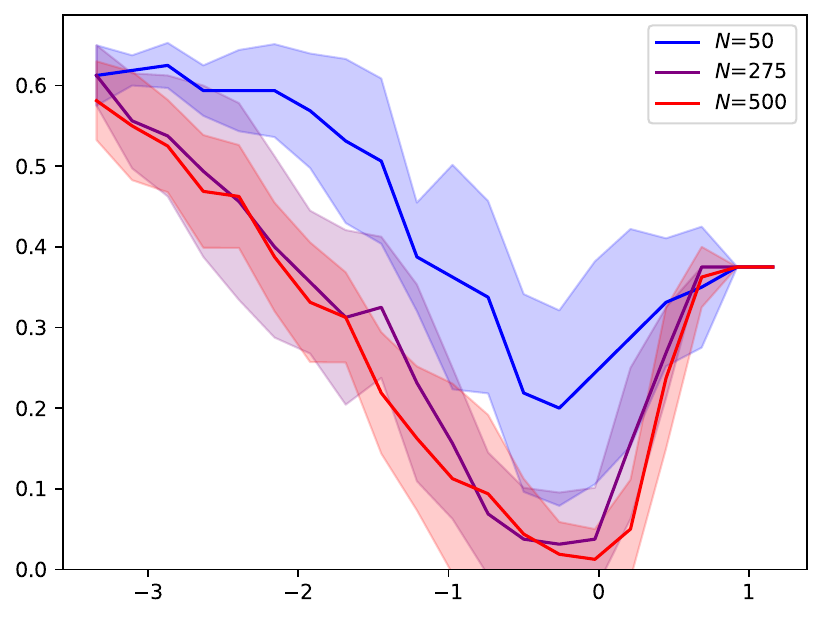}
        \\
        \includegraphics[width=0.3\linewidth]{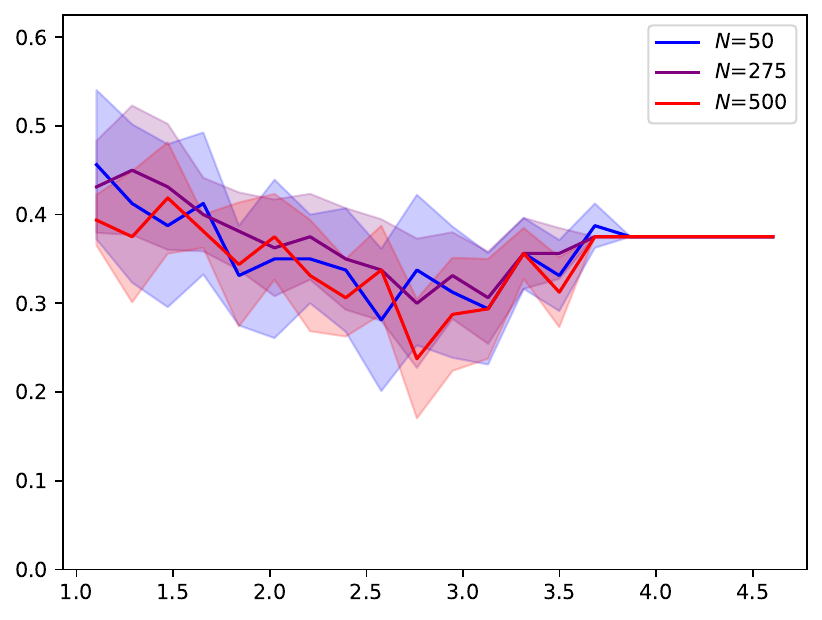}
        &
        \includegraphics[width=0.3\linewidth]{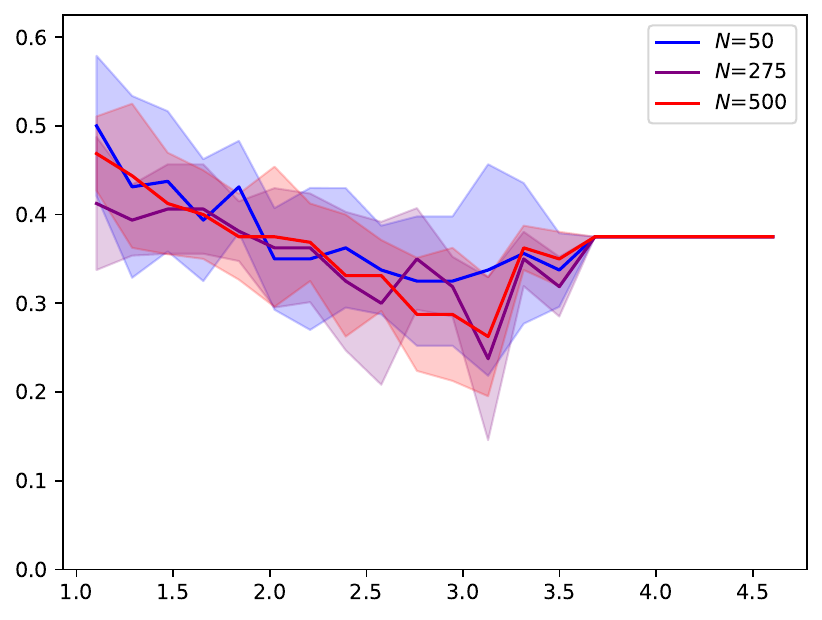}
        &
        \includegraphics[width=0.3\linewidth]{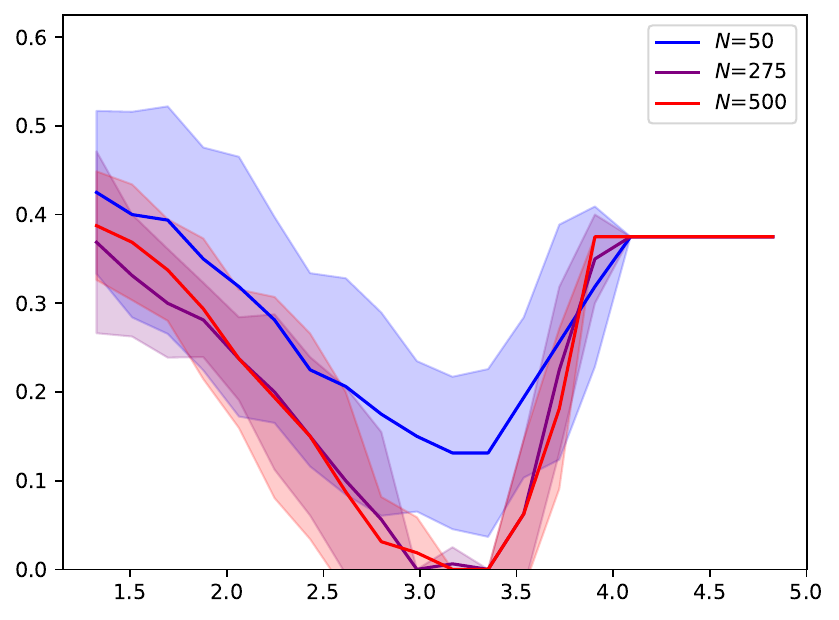}
    \end{tabular}
    \caption{
    Sparse recovery with \(\ell_1\) regularization. Each panel shows the
    fraction of incorrectly estimated matrix entries as a function of
    \(\log(\lambda)\), for increasing sample sizes \(N\). An entry is counted
    as incorrectly estimated when its absolute value exceeds \(10^{-5}\) while
    the corresponding true entry is zero, or fails to exceed this threshold
    while the corresponding true entry is nonzero. The first row corresponds to
    the sharpened Fenchel--Young inverse-gradient-flow loss, while the second
    row shows the quadratic iJKO\(^\star\) baseline.
    }
    \label{fig:recovery-l1}
\end{figure}

\paragraph{Sparse \(\ell_1\)-regularized recovery.}
We take \(\theta^\star\) to be the adjacency matrix of a circular graph:
\[
    \theta^\star_{ij}=1
    \quad\text{if } |i-j|=1 \text{ or } \{i,j\}=\{1,d\},
    \qquad
    \theta^\star_{ij}=0
    \quad\text{otherwise}.
\]
The initial law is
\(\alpha^0=\Nn(2\mathbf 1,\sigma^2\Id)\). We compare three regimes:
\(\sigma=0.1,T=2\), \(\sigma=0.1,T=6\), and \(\sigma=1,T=2\), with
\(\tau=0.1\). Figure~\ref{fig:trajectory-l1} shows the corresponding
population trajectories for several values of \(\sigma\). The recovery plots in
Figure~\ref{fig:recovery-l1} report the fraction of incorrectly estimated
entries after solving the \(\ell_1\)-regularized problem over a grid of
\(\lambda\)'s and sample sizes. We observe that support
recovery improves either when the initial cloud is less concentrated
(\(\sigma=1\)) or when more time snapshots are observed (\(T=6\)). In the
concentrated regime \(\sigma=0.1\), the sharpened Fenchel--Young loss recovers
the support more reliably than the least-squares iJKO\(^\star\) baseline. This
is consistent with the view that iJKO\(^\star\) behaves like a highly localized
first-order surrogate. In concentrated regimes, this localization can harm
sparse identification, whereas the finite-\(r\) sharpened Fenchel--Young loss
retains more global information from the variational transport problem.

\section{Conclusion}

We have developed a gap-based framework for learning potentials from samples
when the observed measure is defined implicitly by a convex variational problem.
The central idea is to replace prediction matching by an optimality-gap loss
that is convex under affine parametrization and remains tied to the geometry of
the forward model. The sharpened Fenchel--Young construction extends this idea
by injecting the observed measure into the nonlinear part of the variational
problem, preserving calibration while improving local curvature and stability.

The abstract stability theorem separates the inverse analysis into three
verifiable components: measurement concentration, forward stability, and local
curvature. We instantiated this modular view for inverse entropic unbalanced
optimal transport and for inverse JKO learning from independent snapshots. In
the latter case, the KL-sharpened JKO objective reduces to an unbalanced
transport problem, which transfers the finite-sample theory and explains why
quadratic iJKO\(^\star\) losses can be understood as local high-sharpening
approximations.

Several questions remain open. The present analysis focuses on
finite-dimensional potential classes; extending the theory to nonparametric
classes would require balancing approximation, curvature, and empirical process
effects. It would also be valuable to sharpen the dependence on dimension and
entropic regularization in the transport bounds, and to develop scalable
algorithms that exploit the improved geometry without solving large inner
transport problems from scratch. More broadly, the same optimality-gap viewpoint
could be useful for inverse variational models beyond optimal transport,
including generative modeling and other distributional inverse problems where
the data are samples from an implicit optimizer.

\section*{Acknowledgments}

The work of F. Andrade and G. Peyr\'e was supported by the French government
under the management of Agence Nationale de la Recherche as part of the
``Investissements d’avenir'' program, reference ANR-19-P3IA-0001 (PRAIRIE 3IA
Institute) and by the European Research Council (ERC project WOLF).

\bibliographystyle{abbrv}
\bibliography{refs}

\appendix

\section{Strong convexity of the exponential functional}

We first record a basic strong convexity estimate for the exponential
functional. Let $\alpha\in\Mm_+(\Xx)$ and $\beta\in\Mm_+(\Yy)$ be finite
positive measures. For functions $f:\Xx\to\RR$, $g:\Yy\to\RR$, and
$c:\Xx\times\Yy\to\RR$, write
\[
    (f\oplus g)(x,y):=f(x)+g(y).
\]

\begin{lem}[Strong convexity of the exponential functional]
\label{lem:scaled_strong_conv_exp}
Let
\[
    \mathcal G_\epsilon(f,g,c)
    :=
    \epsilon
    \left\langle
    \exp\left(\frac{f\oplus g-c}{\epsilon}\right),
    \alpha\otimes\beta
    \right\rangle.
\]
Assume that
\[
    \|f\oplus g-c\|_\infty\leq M,
    \qquad
    \|f'\oplus g'-c'\|_\infty\leq M.
\]
Then
\[
\begin{aligned}
    \mathcal G_\epsilon(f,g,c)
    &\geq
    \mathcal G_\epsilon(f',g',c')
    +
    \left\langle
    \nabla\mathcal G_\epsilon(f',g',c'),
    (f,g,c)-(f',g',c')
    \right\rangle
    \\
    &\qquad
    +
    \frac{1}{2\epsilon}
    \exp\left(-\frac{M}{\epsilon}\right)
    \left\|
    (f-f')\oplus(g-g')-(c-c')
    \right\|_{L^2(\alpha\otimes\beta)}^2.
\end{aligned}
\]
Thus the  exponential functional is
\[
    \epsilon^{-1}\exp(-M/\epsilon)
\]
strongly convex in the variable $f\oplus g-c$ with respect to the
$L^2(\alpha\otimes\beta)$ norm.
\end{lem}

\begin{proof}
Define the interpolation
\[
    f_t:=(1-t)f'+tf,
    \qquad
    g_t:=(1-t)g'+tg,
    \qquad
    c_t:=(1-t)c'+tc,
\]
and set
\[
    F(t):=\mathcal G(f_t,g_t,c_t).
\]
Assume $\epsilon=1$ for now.
Then $F$ is twice differentiable and
\[
\begin{aligned}
    F''(t)
    &=
    \int_{\Xx\times\Yy}
    \exp(f_t\oplus g_t-c_t)
    \left(
    (f-f')\oplus(g-g')-(c-c')
    \right)^2
    d(\alpha\otimes\beta).
\end{aligned}
\]
Since
\[
    \|f_t\oplus g_t-c_t\|_\infty
    \leq
    (1-t)\|f'\oplus g'-c'\|_\infty
    +
    t\|f\oplus g-c\|_\infty
    \leq M,
\]
we have
\[
    F''(t)
    \geq
    e^{-M}
    \int_{\Xx\times\Yy}
    \left(
    (f-f')\oplus(g-g')-(c-c')
    \right)^2
    d(\alpha\otimes\beta).
\]
By Taylor's formula with integral remainder,
\[
    F(1)
    =
    F(0)+F'(0)+\int_0^1(1-t)F''(t)\,dt.
\]
Therefore
\[
\begin{aligned}
    F(1)
    &\geq
    F(0)+F'(0)
    +
    \frac12 e^{-M}
    \int_{\Xx\times\Yy}
    \left(
    (f-f')\oplus(g-g')-(c-c')
    \right)^2
    d(\alpha\otimes\beta),
\end{aligned}
\]
which is the desired strong convexity estimate.

It remains to expand the square. Let
\[
    a:=f-f',
    \qquad
    b:=g-g',
    \qquad
    d:=c-c'.
\]
Then
\[
    (f-f')\oplus(g-g')-(c-c')=a(x)+b(y)-d(x,y).
\]
Expanding and integrating gives
\[
\begin{aligned}
    \int (a+b-d)^2\,d(\alpha\otimes\beta)
    &=
    m_\beta\int a^2\,d\alpha
    +
    m_\alpha\int b^2\,d\beta
    +
    \int d^2\,d(\alpha\otimes\beta)
    \\
    &\quad
    +
    2\left(\int a\,d\alpha\right)
    \left(\int b\,d\beta\right)
    \\
    &\quad
    -
    2\int_{\Xx}a(x)
    \left(\int_{\Yy}d(x,y)\,d\beta(y)\right)
    d\alpha(x)
    \\
    &\quad
    -
    2\int_{\Yy}b(y)
    \left(\int_{\Xx}d(x,y)\,d\alpha(x)\right)
    d\beta(y).
\end{aligned}
\]
The claimed identities follow from the centering assumptions.
Applying this result to 
\[
    \widetilde f:=f/\epsilon,
    \qquad
    \widetilde g:=g/\epsilon,
    \qquad
    \widetilde c:=c/\epsilon,
\]
and multiplying the resulting inequality by $\epsilon$ gives the desired result.
\end{proof}

\section{Concentration of empirically centered feature covariances}

We now prove the concentration estimate used to pass from population
identifiability of centered features to empirical curvature. 
For $i=1,\ldots, n$, let $(X_i,Y_i)\overset{i.i.d.}{\sim} \pi$ where $\pi$ is a probability coupling with marginals $\alpha$ and $\beta$.
Let
\[
    \phi_1,\dots,\phi_S:\Xx\times\Yy\to\RR
\]
be measurable features and write
\[
    \phi(x,y):=(\phi_1(x,y),\dots,\phi_S(x,y))^\top\in\RR^S.
\]

Define the population doubly centered feature map
\[
    \phi^\circ(x,y)
    :=
    \phi(x,y)
    -
    \EE_{X\sim\alpha}\phi(X,y)
    -
    \EE_{Y\sim\beta}\phi(x,Y)
    +
    \EE_{X\sim\alpha,Y\sim\beta}\phi(X,Y).
\]
Assume that
\[
    \|\phi^\circ(x,y)\|_2\leq K
    \qquad
    \text{for all }(x,y)\in\Xx\times\Yy.
\]
Define the population covariance
\[
    \Sigma
    :=
    \EE_{X\sim\alpha,Y\sim\beta}
    \left[
    \phi^\circ(X,Y)\phi^\circ(X,Y)^\top
    \right].
\]

For each $k$, define the empirically doubly centered feature
\[
\begin{aligned}
    \widehat\psi_k(x,y)
    &:=
    \phi_k(x,y)
    -
    \frac1n\sum_{a=1}^n\phi_k(X_a,y)
    -
    \frac1n\sum_{b=1}^n\phi_k(x,Y_b)
    +
    \frac1{n^2}\sum_{a,b=1}^n\phi_k(X_a,Y_b).
\end{aligned}
\]
Let
\[
    \widehat\psi(x,y)
    :=
    \left(
    \widehat\psi_1(x,y),\dots,\widehat\psi_S(x,y)
    \right)^\top.
\]
Finally define the empirical covariance matrix
\[
    \widehat M
    :=
    \frac1{n^2}
    \sum_{i,j=1}^n
    \widehat\psi(X_i,Y_j)\widehat\psi(X_i,Y_j)^\top.
\]

\begin{prop}[Concentration of double-centered covariance]
\label{prop:double_centered_concentration}
There exists a universal constant $C>0$ such that, for every
$\delta\in(0,1)$, with probability at least $1-\delta$,
\[
    \|\widehat M-\Sigma\|_{\mathrm{op}}
    \leq
    C K^2
    \left(
    \sqrt{\frac{\log(2S/\delta)}{n}}
    +
    \frac{\log(2S/\delta)}{n}
    +
    \frac1n
    \right).
\]
Consequently, if
\[
    \lambda_{\min}(\Sigma)\geq \gamma_{\min}
\]
and
\[
    C K^2
    \left(
    \sqrt{\frac{\log(2S/\delta)}{n}}
    +
    \frac{\log(2S/\delta)}{n}
    +
    \frac1n
    \right)
    \leq
    \frac12\gamma_{\min},
\]
then with probability at least $1-\delta$,
\[
    \lambda_{\min}(\widehat M)\geq \frac12\gamma_{\min}
\]
\end{prop}

\begin{proof}
For simplicity of exposition, we first assume that  $\pi$, from which $(X_i,Y_i)$ are sampled, is the independent coupling, so  $Y_i$'s are independent of the $X_i$'s.

Set
\[
    F_{ij}:=\phi^\circ(X_i,Y_j)\in\RR^S.
\]
Let
\[
    P:=I_n-\frac1n\mathbf 1\mathbf 1^\top
\]
be the centering projection. Since the difference between $\phi$ and
$\phi^\circ$ is the sum of a function of $x$, a function of $y$, and a constant,
these terms are removed by empirical double-centering. Therefore
\[
    \widehat\psi(X_i,Y_j)=(PFP)_{ij}.
\]
Thus
\[
    \widehat M
    =
    \frac1{n^2}
    \sum_{i,j=1}^n
    (PFP)_{ij}(PFP)_{ij}^\top.
\]

We first compute the expectation. Since
\[
    \EE_{X\sim\alpha}\phi^\circ(X,y)=0
    \quad\text{for every }y,
\]
and
\[
    \EE_{Y\sim\beta}\phi^\circ(x,Y)=0
    \quad\text{for every }x,
\]
we have
\[
    \EE[F_{ab}F_{cd}^\top]=0
\]
unless $a=c$ and $b=d$. Indeed, if $a\neq c$, then conditioning on all
$Y$-variables and using the centering in the $X$-variable gives zero. Similarly,
if $b\neq d$, conditioning on all $X$-variables and using the centering in the
$Y$-variable gives zero.

Therefore,
\[
\begin{aligned}
    \EE \widehat M
    &=
    \frac1{n^2}
    \sum_{i,j=1}^n
    \sum_{a,b=1}^n
    P_{ia}^2P_{jb}^2\,
    \EE[F_{ab}F_{ab}^\top]
    \\
    &=
    \frac1{n^2}
    \left(
    \sum_{i,a=1}^nP_{ia}^2
    \right)
    \left(
    \sum_{j,b=1}^nP_{jb}^2
    \right)
    \Sigma.
\end{aligned}
\]
Because $P$ is an orthogonal projection of rank $n-1$,
\[
    \sum_{i,a=1}^nP_{ia}^2
    =
    \operatorname{tr}(P^2)
    =
    \operatorname{tr}(P)
    =
    n-1.
\]
Hence
\[
    \EE\widehat M
    =
    \left(1-\frac1n\right)^2\Sigma.
\]
Thus
\[
    \|\EE\widehat M-\Sigma\|_{\mathrm{op}}
    \leq
    \frac{2}{n}\|\Sigma\|_{\mathrm{op}}
    \leq
    \frac{2K^2}{n}.
\]

It remains to control $\widehat M-\EE\widehat M$. Applying
Lemma~\ref{lem:projected_matrix_concentration} below to
\[
    f(x,y):=\phi^\circ(x,y)
\]
gives, with probability at least $1-\delta$,
\[
    \|\widehat M-\EE\widehat M\|_{\mathrm{op}}
    \leq
    C K^2
    \left(
    \sqrt{\frac{\log(2S/\delta)}{n}}
    +
    \frac{\log(2S/\delta)}{n}
    \right).
\]
Combining the two estimates yields the result:
\[
\begin{aligned}
    \|\widehat M-\Sigma\|_{\mathrm{op}}
    &\leq
    \|\widehat M-\EE\widehat M\|_{\mathrm{op}}
    +
    \|\EE\widehat M-\Sigma\|_{\mathrm{op}}
    \\
    &\leq
    C K^2
    \left(
    \sqrt{\frac{\log(2S/\delta)}{n}}
    +
    \frac{\log(2S/\delta)}{n}
    +
    \frac1n
    \right).
\end{aligned}
\]
The eigenvalue claim follows from Weyl's inequality.

If instead $(X_i,Y_i)_{i=1}^n$ are i.i.d. samples from a coupling
$\pi\in\Pi(\alpha,\beta)$, with marginals $\alpha$ and $\beta$, the only change
in this proof is the expectation computation. In that case
$\EE[F_{ab}F_{cd}^\top]$ vanishes whenever some index in the multiset
$\{a,b,c,d\}$ appears only once, by conditioning on all other pairs and using the
corresponding marginal centering. Compared with the independent case, two types
of terms have to be added. First, inside the leading pairing $a=c,\ b=d$, the
diagonal cells $a=b$ are no longer product samples. Second, there are two
additional pair partitions,
\[
    a=b,\ c=d,
    \qquad\text{and}\qquad
    a=d,\ b=c.
\]
All of these terms have total contribution at most $CK^2/n$ in operator norm
after the projection and the normalization by $n^2$. Indeed, using
$\|F_{ab}F_{cd}^\top\|_{\mathrm{op}}\leq K^2$ and $P^2=P$,
\[
\begin{aligned}
    \frac1{n^2}
    \sum_{a=1}^n
    \left(\sum_i P_{ia}^2\right)
    \left(\sum_j P_{ja}^2\right)
    &\leq \frac1n,\\
    \frac1{n^2}
    \sum_{a,c=1}^n
    \left(\sum_i P_{ia}P_{ic}\right)
    \left(\sum_j P_{ja}P_{jc}\right)
    &=
    \frac{\tr(P^2)}{n^2}
    \leq \frac1n,
\end{aligned}
\]
and the crossed pairing is bounded in the same way. Hence the coupled
expectation differs from the independent expectation by a remainder
$R_n$ satisfying $\|R_n\|_{\mathrm{op}}\leq CK^2/n$, which is absorbed by the
existing $n^{-1}$ term in the final bound.
\end{proof}

\begin{lem}[Matrix concentration for a projected two-sample Gram matrix]
\label{lem:projected_matrix_concentration}
Let $X_1,\dots,X_n$ be i.i.d.\ samples from $\alpha$, and let
$Y_1,\dots,Y_n$ be i.i.d.\ samples from $\beta$, independent of the $X_i$'s.
Let
\[
    f:\Xx\times\Yy\to\RR^S
\]
satisfy
\[
    \EE_{X\sim\alpha} f(X,y)=0
    \quad\text{for every }y,
    \qquad
    \EE_{Y\sim\beta} f(x,Y)=0
    \quad\text{for every }x,
\]
and
\[
    \|f(x,y)\|_2\leq K
    \qquad
    \text{for all }(x,y).
\]
Define
\[
\begin{aligned}
    G_{ij}
    &:=
    f(X_i,Y_j)
    -
    \frac1n\sum_{a=1}^n f(X_a,Y_j)
    -
    \frac1n\sum_{b=1}^n f(X_i,Y_b)
    +
    \frac1{n^2}\sum_{a,b=1}^n f(X_a,Y_b),
\end{aligned}
\]
and
\[
    M_f
    :=
    \frac1{n^2}
    \sum_{i,j=1}^nG_{ij}G_{ij}^\top.
\]
Then there exists universal constant $C>0$, for every $\delta\in(0,1)$, with probability at least
$1-\delta$,
\[
    \|M_f-\EE M_f\|_{\mathrm{op}}
    \leq
    C K^2
    \left(
    \sqrt{\frac{\log(2S/\delta)}{n}}
    +
    \frac{\log(2S/\delta)}{n}
    \right).
\]
\end{lem}

\begin{proof}
For simplicity of exposition, we first assume that $\pi$ is the independent coupling, so  $Y_i$'s are independent of the $X_i$'s.

Let $P=I_n-n^{-1}\mathbf 1\mathbf 1^\top$ be the centering projection. Then
\[
    G_{ij}=(PFP)_{ij},
    \qquad
    F_{ij}:=f(X_i,Y_j).
\]
Since $P$ is a contraction and $\|f(x,y)\|_2\leq K$, the triangle inequality
gives the crude pointwise bound
\[
    \|G_{ij}\|_2\leq 4K
    \qquad
    \text{for all }i,j.
\]
Consequently,
\[
    \|G_{ij}G_{ij}^\top\|_{\mathrm{op}}
    \leq
    16K^2.
\]

We use a matrix bounded-differences argument. Reveal the variables in the order
\[
    X_1,\dots,X_n,Y_1,\dots,Y_n
\]
and let $\mathcal F_\ell$
be the corresponding filtration. Let
\[
    D_\ell
    :=
    \EE[M_f\mid\mathcal F_\ell]
    -
    \EE[M_f\mid\mathcal F_{\ell-1}]
\]
be the associated Doob martingale differences. Then
\[
    M_f-\EE M_f=\sum_{\ell=1}^{2n}D_\ell.
\]

We claim that there is a universal constant $C>0$ such that
\[
    \|D_\ell\|_{\mathrm{op}}\leq \frac{C K^2}{n}
    \qquad
    \text{almost surely for every }\ell.
\]
It is enough to consider the effect of replacing one sample, say $X_i$, by an
independent copy $X_i'$. The array $F$ changes only in the $i$th row:
\[
    \Delta F_{ab}
    =
    \mathbf 1_{\{a=i\}}
    \left[
    f(X_i,Y_b)-f(X_i',Y_b)
    \right].
\]
Thus $\|\Delta F_{ib}\|_2\leq 2K$. After applying the projections,
\[
    \Delta G=P(\Delta F)P.
\]
Since the projection spreads the row perturbation across all rows but has
operator norm one, the total contribution of the perturbation to the averaged
Gram matrix is of order $K^2/n$. More explicitly,
\[
\begin{aligned}
    &\left\|
    \frac1{n^2}\sum_{a,b}
    \left[
    (G_{ab}+\Delta G_{ab})(G_{ab}+\Delta G_{ab})^\top
    -
    G_{ab}G_{ab}^\top
    \right]
    \right\|_{\mathrm{op}}
    \\
    &\qquad
    \leq
    \frac1{n^2}
    \sum_{a,b}
    \left(
    2\|G_{ab}\|\,\|\Delta G_{ab}\|
    +
    \|\Delta G_{ab}\|^2
    \right)
    \\
    &\qquad
    \leq
    \frac{C K^2}{n}.
\end{aligned}
\]
The last inequality follows from the row-perturbation structure and the
contraction property of $P$:
\[
    \sum_{a,b}\|\Delta G_{ab}\|^2
    \leq
    \sum_{a,b}\|\Delta F_{ab}\|^2 = \sum_{b}\|\Delta F_{ib}\|^2
    \leq
    CnK^2,
\]
and similarly
\[
    \sum_{a,b}\|G_{ab}\|\,\|\Delta G_{ab}\|\leq Cn^2K^2.
\]
The same argument applies when one replaces a single $Y_j$. Since each martingale
difference $D_\ell$ is a conditional expectation of such a one-sample
replacement difference, the same bound holds:
\[   \|D_\ell\|_{\mathrm{op}}\leq \frac{C K^2}{n}.
\]
Next we bound the quadratic variation. The previous estimate gives
\[
    \|D_\ell^2\|_{\mathrm{op}}
    \leq
    \|D_\ell\|_{\mathrm{op}}^2
    \leq
    \frac{C K^4}{n^2}.
\]
Therefore
\[
    \left\|
    \sum_{\ell=1}^{2n}
    \EE[D_\ell^2\mid \mathcal F_{\ell-1}]
    \right\|_{\mathrm{op}}
    \leq
    \frac{C K^4}{n}.
\]
Set   $R\eqdef C K^2/n$ and $
    \sigma^2\eqdef C K^4/n$.
The matrix Freedman inequality for self-adjoint
martingales~\cite{Tropp2011Freedman}
gives
\[
    \PP\left(
    \left\|\sum_{\ell=1}^{2n}D_\ell\right\|_{\mathrm{op}}\geq t
    \right)
    \leq
    2S
    \exp\left(
    -\frac{t^2/2}{\sigma^2+Rt/3}
    \right).
\]
Substituting the bounds for $R$ and $\sigma^2$ yields
\[
    \PP\left(
    \|M_f-\EE M_f\|_{\mathrm{op}}\geq t
    \right)
    \leq
    2S
    \exp\left[
    -c n
    \min\left\{
    \frac{t^2}{K^4},
    \frac{t}{K^2}
    \right\}
    \right].
\]
In other words, with probability at least $1-\delta$,
\[
    \|M_f-\EE M_f\|_{\mathrm{op}}
    \leq
    C K^2
    \left(
    \sqrt{\frac{log(2S/\delta)}{n}}
    +
    \frac{log(2S/\delta)}{n}
    \right)
\]

If instead $(X_i,Y_i)_{i=1}^n$ are i.i.d. samples from a coupling
$\pi\in\Pi(\alpha,\beta)$, the concentration proof changes only in the
bounded-differences step. One reveals the independent pairs
$(X_i,Y_i)$ rather than the $2n$ marginal variables separately. Replacing one
pair by an independent copy changes the array $F_{ab}=f(X_a,Y_b)$ only in the
union of the corresponding row and column:
\[
    \Delta F_{ab}
    =
    \mathbf 1_{\{a=i\}}
    \bigl[f(X_i,Y_b)-f(X_i',Y_b)\bigr]
    +
    \mathbf 1_{\{b=i\}}
    \bigl[f(X_a,Y_i)-f(X_a,Y_i')\bigr],
\]
with the diagonal entry harmlessly counted twice. This perturbation has at most
$2n-1$ nonzero entries and entrywise norm bounded by $4K$. The same projection
and cross-term estimates above therefore give a one-pair replacement bound
$CK^2/n$. The martingale now has $n$ increments instead of $2n$, so the quadratic variation remains bounded by $CK^4/n$, and the same
Freedman bound follows.
\end{proof}

\section{Curvature of iUOT}\label{app:proof-loc-conv}

This appendix proves the empirical curvature result used in the stability
theory by reducing strong convexity to concentration of identifiable feature
covariances. The proof separates the fully centered, one-sided centered, and
fully curved cases from Assumption~\ref{ass:cost}.

\begin{proof}[Proof of Proposition~\ref{prop:strongConvexity}]
We prove the result by verifying strong convexity of $\widehat J_n$ along the
identifiable feature directions. The proof is divided according to the three
alternatives in Assumption~\ref{ass:cost}. We give the details for cases
(i) and (ii); case (iii) follows similarly, using the local strong convexity of
both marginal dual terms. Throughout the proof, write
\[
    \widehat\nu_1:=\widehat\nu_1^n,
    \qquad
    \widehat\nu_2:=\widehat\nu_2^n,
    \qquad
    m_i:=m_{\nu_i^\star}.
\]

\paragraph{Empirical centering.}

Define the empirical marginal averages of the feature map by
\begin{align}
    \widehat\phi^{(1)}(x)
    &:=
    \frac{1}{m_2}
    \int_{\Yy}\phi(x,y)\,d\widehat\nu_2(y),
    \\
    \widehat\phi^{(2)}(y)
    &:=
    \frac{1}{m_1}
    \int_{\Xx}\phi(x,y)\,d\widehat\nu_1(x),
    \\
    \widehat\phi^{(12)}
    &:=
    \frac{1}{m_1m_2}
    \int_{\Xx\times\Yy}
    \phi(x,y)\,
    d\widehat\nu_1(x)d\widehat\nu_2(y).
\end{align}
Here $\phi=(\phi_1,\dots,\phi_S)$ is vector-valued. Hence each of
$\widehat\phi^{(1)},\widehat\phi^{(2)},\widehat\phi^{(12)}$ is also
$\RR^S$-valued.

For the affine parametrization
\[
    V_\theta(x,y)
    =
    \phi_0(x,y)+\theta^\top\phi(x,y),
\]
define the empirically centered features
\[
    \widehat\psi^{(0)}(x,y)
    :=
    \phi(x,y)
    -
    \widehat\phi^{(1)}(x)
    -
    \widehat\phi^{(2)}(y)
    +
    \widehat\phi^{(12)}.
\]
Then decompose
\[
    V_\theta(x,y)
    =
    V_\theta^{(0)}(x,y)
    +
    V_\theta^{(1)}(x)
    +
    V_\theta^{(2)}(y),
\]
where
\begin{align}
    V_\theta^{(0)}(x,y)
    &:=
    \phi_0(x,y)+\theta^\top\widehat\psi^{(0)}(x,y),
    \\
    V_\theta^{(1)}(x)
    &:=
    \theta^\top\widehat\phi^{(1)}(x),
    \\
    V_\theta^{(2)}(y)
    &:=
    \theta^\top\left(
    \widehat\phi^{(2)}(y)-\widehat\phi^{(12)}
    \right).
\end{align}
This decomposition is useful because
\[
    \int_{\Yy}\widehat\psi^{(0)}(x,y)\,d\widehat\nu_2(y)=0
    \quad\text{for every }x,
\]
and
\[
    \int_{\Xx}\widehat\psi^{(0)}(x,y)\,d\widehat\nu_1(x)=0
    \quad\text{for every }y.
\]

\paragraph{Dual representation.}

Recall that
\[
    \widehat\Omega_n(\mu)
    =
    \epsilon\KL(\mu\mid\widehat\nu_1\otimes\widehat\nu_2)
    +
    D_{\varphi_1}(\mu_1\mid\widehat\nu_1)
    +
    D_{\varphi_2}(\mu_2\mid\widehat\nu_2).
\]
By convex duality for entropic unbalanced optimal transport,
\begin{align*}
    &\inf_{\mu\in\Mm_+(\Xx\times\Yy)}
    \left\{
    \langle V_\theta,\mu\rangle+\widehat\Omega_n(\mu)
    \right\}
    \\
    &\qquad
    =
    \sup_{f,g}
    \Bigg\{
    -
    \int_{\Xx}\varphi_1^\ast(-f(x))\,d\widehat\nu_1(x)
    \\
    &\qquad\quad
    -
    \int_{\Yy}\varphi_2^\ast(-g(y))\,d\widehat\nu_2(y)
    \\
    &\qquad\quad
    -
    \epsilon
    \int_{\Xx\times\Yy}
    \exp\left(
    \frac{f(x)+g(y)-V_\theta(x,y)}{\epsilon}
    \right)
    d\widehat\nu_1(x)d\widehat\nu_2(y)
    \Bigg\}.
\end{align*}
Consequently, the empirical Fenchel--Young objective can be written as
\begin{align}
    \widehat J_n(\theta)
    =
    \inf_{f,g}
    \Bigg\{
    &\langle V_\theta,\widehat\mu_n\rangle
    +
    \int_{\Xx}\varphi_1^\ast(-f(x))\,d\widehat\nu_1(x)
    +
    \int_{\Yy}\varphi_2^\ast(-g(y))\,d\widehat\nu_2(y)
    \nonumber\\
    &+
    \epsilon
    \int_{\Xx\times\Yy}
    \exp\left(
    \frac{f(x)+g(y)-V_\theta(x,y)}{\epsilon}
    \right)
    d\widehat\nu_1(x)d\widehat\nu_2(y)
    \Bigg\}.
\end{align}

Using the decomposition of $V_\theta$, and making the change of variables
\[
    f\mapsto f+V_\theta^{(1)},
    \qquad
    g\mapsto g+V_\theta^{(2)},
\]
we obtain
\begin{align}\label{eq:inter_Jn}
    \widehat J_n(\theta)
    =
    \inf_{f,g}
    \Bigg\{
    &\langle V_\theta,\widehat\mu_n\rangle
    +
    \int_{\Xx}
    \varphi_1^\ast(-f(x)-V_\theta^{(1)}(x))
    d\widehat\nu_1(x)
    \nonumber\\
    &+
    \int_{\Yy}
    \varphi_2^\ast(-g(y)-V_\theta^{(2)}(y))
    d\widehat\nu_2(y)
    \nonumber\\
    &+
    \epsilon
    \int_{\Xx\times\Yy}
    \exp\left(
    \frac{f(x)+g(y)-V_\theta^{(0)}(x,y)}{\epsilon}
    \right)
    d\widehat\nu_1(x)d\widehat\nu_2(y)
    \Bigg\}.
\end{align}

Because the final term in \eqref{eq:inter_Jn}  invariant under the transformation
\[
    f\mapsto f+\lambda,
    \qquad
    g\mapsto g-\lambda,
\]
we fix a gauge and introduce a free parameter $\lambda\in\RR$ in the other terms. Specifically, we restrict to
\[
    \mathcal P_0
    :=
    \left\{
    (f,g):
    \|f\|_\infty,\|g\|_\infty\leq M,
    \quad
    \int_{\Xx}f\,d\widehat\nu_1=0
    \right\}.
\]
The bound $\|f\|_\infty,\|g\|_\infty\leq M$ is justified by the standard
a priori bounds for the dual potentials on compact domains, for
$\|\theta\|\leq B$~\cite{poon_UOT_complexity}. The constant $M$ depends on
$B$, the feature bound, the masses, and the local properties of
$\varphi_1^\ast,\varphi_2^\ast$.

Equivalently,
\begin{align}
    \widehat J_n(\theta)
    =
    \inf_{(f,g)\in\mathcal P_0,\lambda\in\RR}
    \left\{
    \mathcal G_0(f,g,\theta)
    +
    \mathcal G_1(f,g,\theta,\lambda)
    \right\},
\end{align}
where
\begin{align}
    \mathcal G_0(f,g,\theta)
    &:=
    \langle V_\theta^{(0)},\widehat\mu_n\rangle
    +
    \epsilon
    \int_{\Xx\times\Yy}
    \exp\left(
    \frac{f(x)+g(y)-V_\theta^{(0)}(x,y)}{\epsilon}
    \right)
    d\widehat\nu_1(x)d\widehat\nu_2(y),
    \\
    \mathcal G_1(f,g,\theta,\lambda)
    &:=
    \langle V_\theta^{(1)}\oplus V_\theta^{(2)},\widehat\mu_n\rangle
    \nonumber\\
    &\quad+
    \int_{\Xx}
    \varphi_1^\ast(-f(x)-\lambda-V_\theta^{(1)}(x))
    d\widehat\nu_1(x)
    \nonumber\\
    &\quad+
    \int_{\Yy}
    \varphi_2^\ast(-g(y)+\lambda-V_\theta^{(2)}(y))
    d\widehat\nu_2(y).
\end{align}

\paragraph{Case (i): curvature in the fully centered cost direction.}

We first assume Assumption~\ref{ass:cost}(i). The term $\mathcal G_1$ is convex
in $(f,g,\theta,\lambda)$. The term $\mathcal G_0$ is strongly convex in the
quantity
\[
    f\oplus g - V_\theta^{(0)}.
\]
By Lemma~\ref{lem:scaled_strong_conv_exp}, for all feasible
$(f,g,\lambda)$ and $(f',g',\lambda')$, and for
\[
    \theta_t=t\theta+(1-t)\theta',
    \quad
    f_t=tf+(1-t)f',
    \quad
    g_t=tg+(1-t)g',
    \quad
    \lambda_t=t\lambda+(1-t)\lambda',
\]
one has
\begin{align}
    \mathcal G_0(f_t,g_t,\theta_t)
    &\leq
    t\mathcal G_0(f,g,\theta)
    +(1-t)\mathcal G_0(f',g',\theta')
    \nonumber\\
    &\quad
    -
    \frac{\gamma_0}{2}t(1-t)
    \left\|
    (f-f')\oplus(g-g')
    -
    \left(V_\theta^{(0)}-V_{\theta'}^{(0)}\right)
    \right\|^2_{L^2(\widehat\nu_1\otimes\widehat\nu_2)}.
\end{align}
By Lemma~\ref{lem:scaled_strong_conv_exp}, one may take
\[
    \gamma_0
    =
    \epsilon^{-1}\exp(-M/\epsilon),
\]
up to universal constants.

Since
\[
    V_\theta^{(0)}-V_{\theta'}^{(0)}
    =
    (\theta-\theta')^\top\widehat\psi^{(0)},
\]
and since $\widehat\psi^{(0)}$ is empirically centered in both variables, it is
orthogonal in $L^2(\widehat\nu_1\otimes\widehat\nu_2)$ to all additive functions
of the form $a(x)+b(y)$. Therefore,
\begin{align}
    &\left\|
    (f-f')\oplus(g-g')
    -
    (\theta-\theta')^\top\widehat\psi^{(0)}
    \right\|^2_{L^2(\widehat\nu_1\otimes\widehat\nu_2)}
    \nonumber\\
    &\qquad
    =
    \|(f-f')\oplus(g-g')\|^2_{L^2(\widehat\nu_1\otimes\widehat\nu_2)}
    +
    \left\|
    (\theta-\theta')^\top\widehat\psi^{(0)}
    \right\|^2_{L^2(\widehat\nu_1\otimes\widehat\nu_2)}
    \nonumber\\
    &\qquad
    \geq
    \left\|
    (\theta-\theta')^\top\widehat\psi^{(0)}
    \right\|^2_{L^2(\widehat\nu_1\otimes\widehat\nu_2)}.
\end{align}
Using also the convexity of $\mathcal G_1$, we obtain
\begin{align}
    \widehat J_n(\theta_t)
    &\leq
    t\mathcal G_0(f,g,\theta)
    +(1-t)\mathcal G_0(f',g',\theta')
    +
    t\mathcal G_1(f,g,\theta,\lambda)
    +(1-t)\mathcal G_1(f',g',\theta',\lambda')
    \nonumber\\
    &\quad
    -
    \frac{\gamma_0}{2}t(1-t)
    \left\|
    (\theta-\theta')^\top\widehat\psi^{(0)}
    \right\|^2_{L^2(\widehat\nu_1\otimes\widehat\nu_2)}.
\end{align}
Taking the infimum over both triples
$(f,g,\lambda)$ and $(f',g',\lambda')$ gives
\begin{align}
    \widehat J_n(\theta_t)
    \leq
    t\widehat J_n(\theta)
    +(1-t)\widehat J_n(\theta')
    -
    \frac{\gamma_0}{2}t(1-t)
    \left\|
    (\theta-\theta')^\top\widehat\psi^{(0)}
    \right\|^2_{L^2(\widehat\nu_1\otimes\widehat\nu_2)}.
\end{align}

Now define the empirical centered feature covariance
\[
    \widehat M_0
    :=
    \int
    \widehat\psi^{(0)}(x,y)\widehat\psi^{(0)}(x,y)^\top\,
    d\widehat\nu_1(x)d\widehat\nu_2(y).
\]
Then
\[
    \left\|
    (\theta-\theta')^\top\widehat\psi^{(0)}
    \right\|^2_{L^2(\widehat\nu_1\otimes\widehat\nu_2)}
    =
    (\theta-\theta')^\top
    \widehat M_0
    (\theta-\theta').
\]
By Assumption~\ref{ass:cost}(i), the population matrix
\[
    M_0
    :=
    \int
    \bar\phi^{(0)}(x,y)\bar\phi^{(0)}(x,y)^\top\,
    d\bar\nu_1^\star(x)d\bar\nu_2^\star(y)
\]
has smallest eigenvalue $\alpha_{\min}$.
By Proposition~\ref{prop:double_centered_concentration},
\[
    \|\widehat M_0-M_0\|_{\mathrm{op}}
    \leq
    \frac12\alpha_{\min},
\]
with probability at least $1-\delta$. Hence
$\lambda_{\min}(\widehat M_0)
    \geq
    \frac12\alpha_{\min}$.
Therefore,
\begin{align}
    \widehat J_n(\theta_t)
    \leq
    t\widehat J_n(\theta)
    +(1-t)\widehat J_n(\theta')
    -
    \frac{\gamma_0\alpha_{\min}}{4}
    t(1-t)\|\theta-\theta'\|^2.
\end{align}
This proves strong convexity in case (i), with
\[
    \alpha
    \gtrsim
    \gamma_0\alpha_{\min}
    \gtrsim
    \alpha_{\min}\exp(-C B/\epsilon).
\]

\paragraph{Case (ii): one marginal divergence has curvature.}

We next assume Assumption~\ref{ass:cost}(ii), namely that
$\varphi_2^\ast$ is locally strongly convex and that the one-sided centered
feature covariance is nondegenerate. Define the one-sided empirical centering
\[
    \widehat\psi^{(1)}(x,y)
    :=
    \phi(x,y)-\widehat\phi^{(1)}(x),
    \qquad
    \widehat\phi^{(1)}(x)
    =
    \frac1{m_2}\int\phi(x,y)\,d\widehat\nu_2(y).
\]
Decompose
\[
    V_\theta(x,y)
    =
    V_\theta^{Y}(x,y)+V_\theta^{X}(x),
\]
where
\[
    V_\theta^{Y}(x,y)
    :=
    \phi_0(x,y)+\theta^\top\widehat\psi^{(1)}(x,y),
    \qquad
    V_\theta^{X}(x)
    :=
    \theta^\top\widehat\phi^{(1)}(x).
\]
By construction,
\[
    \int_{\Yy}
    \left(
    V_\theta^{Y}(x,y)-\phi_0(x,y)
    \right)
    d\widehat\nu_2(y)
    =
    0
    \qquad
    \text{for every }x.
\]

Using the same dual representation and shifting the $x$-only term into the
first dual potential, we can write
\begin{align}
    \widehat J_n(\theta)
    =
    \inf_{(f,g)\in\mathcal P_0,\lambda\in\RR}
    \Bigg\{
    &\langle V_\theta,\widehat\mu_n\rangle
    +
    \int_{\Xx}
    \varphi_1^\ast(-f(x)-\lambda-V_\theta^{X}(x))
    d\widehat\nu_1(x)
    \nonumber\\
    &+
    \int_{\Yy}
    \varphi_2^\ast(-g(y)+\lambda)
    d\widehat\nu_2(y)
    \nonumber\\
    &+
    \epsilon
    \int_{\Xx\times\Yy}
    \exp\left(
    \frac{f(x)+g(y)-V_\theta^{Y}(x,y)}{\epsilon}
    \right)
    d\widehat\nu_1(x)d\widehat\nu_2(y)
    \Bigg\},
\end{align}
where now we fix the gauge by taking
\[
    \mathcal P_0
    :=
    \left\{
    (f,g):
    \|f\|_\infty,\|g\|_\infty\leq M,
    \quad
    \int_{\Yy}g\,d\widehat\nu_2=0
    \right\}.
\]

Define
\begin{align}
    \mathcal G_1(f,\theta,\lambda)
    &:=
    \langle V_\theta,\widehat\mu_n\rangle
    +
    \int_{\Xx}
    \varphi_1^\ast(-f(x)-\lambda-V_\theta^{X}(x))
    d\widehat\nu_1(x),
    \\
    \mathcal G_2(g,\lambda)
    &:=
    \int_{\Yy}
    \varphi_2^\ast(-g(y)+\lambda)
    d\widehat\nu_2(y),
    \\
    \mathcal G_3(f,g,\theta)
    &:=
    \epsilon
    \int_{\Xx\times\Yy}
    \exp\left(
    \frac{f(x)+g(y)-V_\theta^{Y}(x,y)}{\epsilon}
    \right)
    d\widehat\nu_1(x)d\widehat\nu_2(y).
\end{align}
The term $\mathcal G_1$ is convex. Since $\varphi_2^\ast$ is locally strongly
convex on the relevant range, $\mathcal G_2$ is strongly convex in $g-\lambda$.
After fixing the gauge $\int g\,d\widehat\nu_2=0$, this yields a constant
$\gamma_2>0$ such that
\[
    \mathcal G_2(g_t,\lambda_t)
    \leq
    t\mathcal G_2(g,\lambda)
    +(1-t)\mathcal G_2(g',\lambda')
    -
    \frac{\gamma_2}{2}t(1-t)
    \|g-g'\|^2_{L^2(\widehat\nu_2)}.
\]
Moreover, by Lemma~\ref{lem:scaled_strong_conv_exp}, $\mathcal G_3$ is strongly convex
in
\[
    u:=f\oplus g - V_\theta^{Y}.
\]
Thus, for some
\[
    \gamma_3
    \gtrsim
    \epsilon^{-1}\exp(-C M'(B)/\epsilon),
\]
again up to the normalization of Lemma~\ref{lem:scaled_strong_conv_exp},
\[
    \mathcal G_3(f_t,g_t,\theta_t)
    \leq
    t\mathcal G_3(f,g,\theta)
    +(1-t)\mathcal G_3(f',g',\theta')
    -
    \frac{\gamma_3}{2}t(1-t)
    \|u-u'\|^2_{L^2(\widehat\nu_1\otimes\widehat\nu_2)}.
\]
Combining these inequalities and taking infima gives
\begin{align}
\label{eq:strong_conv_interm_clean}
    \widehat J_n(\theta_t)
    &\leq
    t\widehat J_n(\theta)
    +(1-t)\widehat J_n(\theta')
    \nonumber\\
    &\quad
    -
    \frac{\gamma_2}{2}t(1-t)
    \|g-g'\|^2_{L^2(\widehat\nu_2)}
    -
    \frac{\gamma_3}{2}t(1-t)
    \|u-u'\|^2_{L^2(\widehat\nu_1\otimes\widehat\nu_2)}.
\end{align}
Here the inequality is first written for arbitrary feasible competitors and then
passed to the infimum in the same way as in case (i).

We now extract curvature in $\theta$. Let
\[
    h:=g-g',
    \qquad
    q(x,y):=(\theta-\theta')^\top\widehat\psi^{(1)}(x,y).
\]
Then
\[
    u-u'
    =
    (f-f')\oplus(g-g')-q.
\]
Since $q$ is centered in the $y$-variable,
\[
    \int_{\Yy}q(x,y)\,d\widehat\nu_2(y)=0
    \qquad
    \text{for every }x.
\]
Therefore the $x$-only term $f-f'$ is orthogonal to $q$ in
$L^2(\widehat\nu_1\otimes\widehat\nu_2)$. Hence
\begin{align}
    \|u-u'\|^2_{L^2(\widehat\nu_1\otimes\widehat\nu_2)}
    &=
    \|(f-f')\oplus h-q\|^2
    \nonumber\\
    &\geq
    \|h-q\|^2_{L^2(\widehat\nu_1\otimes\widehat\nu_2)}.
\end{align}
Moreover,
\[
    \|h-q\|^2
    =
    \|h\|^2_{L^2(\widehat\nu_2)}
    +
    \|q\|^2_{L^2(\widehat\nu_1\otimes\widehat\nu_2)}
    -
    2\langle h,q\rangle.
\]
For any $\rho>0$, Young's inequality gives
\[
    2|\langle h,q\rangle|
    \leq
    \rho\|h\|^2
    +
    \rho^{-1}\|q\|^2.
\]
Optimizing the resulting lower bound in combination with the separate
$\gamma_2\|h\|^2$ term yields the elementary inequality
\[
    \gamma_2\|h\|^2
    +
    \gamma_3\|h-q\|^2
    \geq
    \frac{\gamma_2\gamma_3}{\gamma_2+\gamma_3}
    \|q\|^2.
\]
Substituting this into~\eqref{eq:strong_conv_interm_clean}, we obtain
\[
    \widehat J_n(\theta_t)
    \leq
    t\widehat J_n(\theta)
    +(1-t)\widehat J_n(\theta')
    -
    \frac{t(1-t)}{2}
    \frac{\gamma_2\gamma_3}{\gamma_2+\gamma_3}
    \left\|
    (\theta-\theta')^\top\widehat\psi^{(1)}
    \right\|^2_{L^2(\widehat\nu_1\otimes\widehat\nu_2)}.
\]

Define the empirical one-sided centered covariance matrix
\[
    \widehat M_1
    :=
    \int
    \widehat\psi^{(1)}(x,y)\widehat\psi^{(1)}(x,y)^\top
    d\widehat\nu_1(x)d\widehat\nu_2(y).
\]
Then
\[
    \left\|
    (\theta-\theta')^\top\widehat\psi^{(1)}
    \right\|^2_{L^2(\widehat\nu_1\otimes\widehat\nu_2)}
    =
    (\theta-\theta')^\top
    \widehat M_1
    (\theta-\theta').
\]
By Assumption~\ref{ass:cost}(ii) and the concentration argument of
Proposition~\ref{prop:double_centered_concentration} (the proof of this one sided concentration is analogous and hence omitted),
\[
    \lambda_{\min}(\widehat M_1)
    \geq
    \frac12\alpha_{\min}
\]
with probability at least $1-\delta$. Therefore
\[
    \widehat J_n(\theta_t)
    \leq
    t\widehat J_n(\theta)
    +(1-t)\widehat J_n(\theta')
    -
    \frac{\alpha_{\min}}{4}
    \frac{\gamma_2\gamma_3}{\gamma_2+\gamma_3}
    t(1-t)\|\theta-\theta'\|^2.
\]
Since $\gamma_2$ is bounded below on the relevant compact range and
\[
    \gamma_3
    \gtrsim
    \exp(-C B/\epsilon),
\]
this gives
\[
    \alpha
    \gtrsim
    \alpha_{\min}\exp(-C B/\epsilon).
\]

\paragraph{Case (iii).}

If both $\varphi_1^\ast$ and $\varphi_2^\ast$ are locally strongly convex, then
the marginal dual terms provide curvature in both additive directions. In this
case one does not need to quotient out either marginal component. Repeating the
argument above without centering the feature map gives curvature controlled by
the empirical covariance
\[
    \widehat M
    :=
    \int
    \phi(x,y)\phi(x,y)^\top\,
    d\widehat\nu_1(x)d\widehat\nu_2(y).
\]
By Assumption~\ref{ass:cost}(iii) and the same concentration argument,
\[
    \lambda_{\min}(\widehat M)\geq \frac12\alpha_{\min}
\]
with probability at least $1-\delta$. Combining the exponential curvature of
the entropic term with the local curvature of both marginal dual terms again
yields
\[
    \alpha
    \geq
    C_1\alpha_{\min}\exp(-C_2B/\epsilon).
\]

Combining the three cases proves the proposition.
\end{proof}
\end{document}